\documentclass{article}
\usepackage[inline]{enumitem}

\usepackage{microtype}
\usepackage{graphicx}
\usepackage{epstopdf}
\usepackage{subfigure}
\usepackage{booktabs} 
\usepackage{extarrows}

\usepackage{hyperref}
\usepackage{url}
\usepackage{mathtools,thmtools}
\usepackage{nicefrac}
\usepackage{xfrac}
\usepackage{amsmath,amsfonts,bm}
\usepackage{amssymb}
\usepackage{soul}
\usepackage{booktabs} 
\usepackage{extarrows}
\usepackage{multirow}
\usepackage{tabularx}
\usepackage[hang,flushmargin]{footmisc}

\usepackage{amssymb}
\usepackage{relsize}
\usepackage{enumitem}
\usepackage{multirow}
\usepackage{array}
\usepackage{subfigure}
\usepackage{color}
\usepackage{xfrac}

\usepackage{tikz}

\newcommand{\Arrow}[1]{%
\parbox{#1}{\tikz{\draw[->](0,0)--(#1,0);}}
}
\newcommand{\shortarrow}{\Arrow{1.5mm}}
  
\newcommand{\T}{m}

\newcommand{\pref}[1]{Eq.~(\ref{#1})}

\newcommand{\method}{IT-MTL}

\definecolor{myred}{HTML}{e53935}
\definecolor{myblue}{HTML}{0277bd}

\definecolor{modelblue}{HTML}{1034A6}
\definecolor{urlorange}{HTML}{e68a00}

\newcommand{\total}{\text{\tiny total}}

\usepackage[export]{adjustbox}

\usepackage{amsmath,amsfonts,bm}

\def\@fnsymbol#1{\ensuremath{\ifcase#1\or *\or \dagger\or \ddagger\or
   \mathsection\or \mathparagraph\or \|\or **\or \dagger\dagger
   \or \ddagger\ddagger \else\@ctrerr\fi}}
\newcommand{\ssymbol}[1]{^{\@fnsymbol{#1}}}

\def\Tableref#1{Table~\ref{#1}}

\def\Figref#1{Figure~\ref{#1}}

\def\Secref#1{Section~\ref{#1}}

\def\eqref#1{equation~\ref{#1}}

\def\Algref#1{Algorithm~\ref{#1}}

\def\1{\bm{1}}

\DeclareMathAlphabet{\mathsfit}{\encodingdefault}{\sfdefault}{m}{sl}
\SetMathAlphabet{\mathsfit}{bold}{\encodingdefault}{\sfdefault}{bx}{n}

\DeclareMathOperator*{\argmax}{arg\,max}

\newcommand{\thet}{\bm{\theta}}

\graphicspath{{figures/}}

\hypersetup{
    linkcolor=blue,
}

\usepackage[accepted]{icml2021}

\icmltitlerunning{Measuring and Harnessing Transference in Multi-Task Learning}

\begin{document}
\onecolumn
\thispagestyle{simple}

\vspace*{2cm}
\begin{list}{}{%
  \leftmargin=.15\textwidth
  \rightmargin=.15\textwidth
  \listparindent=\parindent
  \itemindent=\parindent
  \itemsep=0pt
  \parsep=0pt}
\item\relax

A short note to the reader,\\

Many of the ideas and concepts initially presented in this work are being divided into two distinct papers. The first, \textbf{Efficiently Identifying Task Groupings in Multi-Task Learning} more rigorously and comprehensively develops the concept of looking at how the gradient of one task would affect the loss of another to identify which tasks should train together. The second is currently being written and will focus on using inter-task affinity at micro-level to both analyze and modify the optimization dynamics of multi-task neural networks. We direct researchers endeavoring to cite this work to \textbf{Efficiently Identifying Task Groupings in Multi-Task Learning}.\\

One especially salient change is renaming the measure ``Transference'' to ``Inter-Task Affinity''. The word ``Transference'' conveys a unique connotation in the sciences, especially in the field of psychotherapy. We refer interested readers to the \href{https://en.wikipedia.org/wiki/Transference}{Transference Wikipedia page} for additional information.\\

Our best regards,\\
The Authors
\end{list}
\vfill 
\newpage

\twocolumn[
\icmltitle{Measuring and Harnessing Transference in Multi-Task Learning}

\begin{icmlauthorlist}
\icmlauthor{Christopher Fifty}{gbrain}
\icmlauthor{Ehsan Amid}{gresearch}
\icmlauthor{Zhe Zhao}{gbrain}
\icmlauthor{Tianhe Yu}{gbrain,stanford}
\icmlauthor{Rohan Anil}{gbrain}
\icmlauthor{Chelsea Finn}{gbrain,stanford}
\end{icmlauthorlist}

\icmlaffiliation{gbrain}{Google Brain}
\icmlaffiliation{gresearch}{Google Research}
\icmlaffiliation{stanford}{Stanford University}

\icmlcorrespondingauthor{Christopher Fifty}{cfifty@google.com}

\icmlkeywords{Machine Learning, ICML}
\vskip 0.3in
]

\printAffiliationsAndNotice{} 

\begin{abstract}
Multi-task learning can leverage information learned by one task to benefit the training of other tasks. Despite this capacity, na\"ive formulations often degrade performance, and in particular, efficiently identifying the tasks that would benefit from co-training remains a challenging design question. In this paper, we analyze the dynamics of information transfer, or \emph{transference}, across tasks throughout training. Specifically, we develop a similarity measure that can quantify transference among tasks and use this quantity to both better understand the optimization dynamics of multi-task learning as well as improve overall learning performance. In the latter case, we propose two methods to leverage transference. The first operates at a macro-level by selecting which tasks should train together while the second functions at a micro-level by modifying the gradients at each training step. We find these methods can lead to improvement over prior work on five supervised multi-task learning benchmarks and one multi-task reinforcement learning paradigm.
\end{abstract}

\section{Introduction}
Deciding if two or more objectives should be trained together in a multi-task model, as well as choosing how that model's parameters should be shared, is an inherently complex issue often left to human experts~\citep{zhang2017survey}. However, a human's understanding of similarity is motivated by their intuition and experience rather than a prescient knowledge of the underlying structures learned by a neural network. To further complicate matters, the benefit or detriment induced from co-training relies on many non-trivial decisions including, but not limited to, dataset characteristics, model architecture, hyperparameters, capacity, and convergence~\citep{wu2020understanding, branched, standley2019tasks, sun2019adashare}. As a result, a quantifiable measure which conveys the effect of information transfer in a neural network would be valuable to practitioners and researchers alike to better construct or understand multi-task learning paradigms~\citep{baxter2000, ben2003exploiting}. 

The training dynamics specific to multitask neural networks, namely cross-task interactions at the shared parameters~\citep{modulation}, are difficult to predict and only fully manifest at the completion of training. Given the cost, both with regards to time and resources, of fully training a deep neural network, an exhaustive search over the $2^\T - 1 $ possible combinations of $\T$ tasks to determine ideal task groupings can be infeasible. This search is only complicated by the irreproducibility inherent in traversing a loss landscape with high curvature; an effect which appears especially pronounced in multi-task learning paradigms~\citep{pcgrad, standley2019tasks}.

In this paper, we aim to take a step towards quantifying \emph{transference}, or the dynamics of information transfer, and understanding its effect on multi-task training efficiency. As both the input data and state of model convergence are fundamental to transference~\citep{wu2020understanding}, we develop a parameter-free approach to measure this effect at a per-minibatch level of granularity. Moreover, our quantity makes no assumptions regarding model architecture, and is applicable to any paradigm in which shared parameters are updated with respect to multiple task losses.

By analyzing multi-task training dynamics through the lens of transference, we find excluding certain task gradients from the multi-task gradient update for select minibatches can improve learning efficiency. Our analysis suggests this is due to large variation in loss landscapes for different tasks as illustrated in \Figref{fig:1dlosslandscape} and \Figref{fig:2dlosslandscape}. Building on this observation, we propose two methods to utilize transference in multi-task learning algorithms -- to choose which tasks to train together as well as determining which gradients to apply at each minibatch. Our experiments indicate the former can identify promising task groupings, while the latter can improve learning performance over prior methods. 

In summary, our main contributions are three-fold: we (i) propose the first measure (to our knowledge) which quantifies information transfer among tasks in multi-task learning; (ii) demonstrate how transference can be used as a heuristic to select task groupings; (iii) present a method which leverages minibatch-level transference to augment network performance.

\section{Related Work}
\textbf{Multi-Task Formulation.} The most prevalent formulation of MTL is \textit{hard parameter sharing} of hidden layers~\citep{ruder_overview, caruana1993multitask}. In this design, a subset of the hidden layers are typically shared among all tasks, and task-specific layers are stacked on top of the shared base to output a prediction value. Each task is assigned a weight, and the loss of the entire model is a linear combination of each task's loss multiplied by its respective loss weight. This particular design enables parameter efficiency by sharing hidden layers across tasks, reduces overfitting, and can facilitate transfer learning effects among tasks~\citep{ruder_overview, baxter2000, zhang2017survey}.

\textbf{Information Sharing.} Prevailing wisdom suggests tasks which are similar or share a similar underlying structure may benefit from co-training in a multi-task system~\citep{caruana1993multitask, caruana1998multitask}. A plethora of multi-task methods addressing \textit{what to share} have been developed, such as Neural Architecture Search~\citep{guo2020learning, sun2019adashare, branched, rusu2016progressive, Huang_2018, Lu_2017} and Soft-Parameter Sharing~\citep{cross_stitch, duong2015low, yang2016trace}, to improve multi-task performance. Though our measure of transference is complementary with these methods, we direct our focus towards which tasks should be trained together rather than architecture modifications to maximize the benefits of co-training. 

While deciding which tasks to train together has traditionally been addressed with costly cross-validation techniques or high variance human intuition, recent advances have developed increasingly efficient algorithms to assess co-training performance. \citet{swirszcz2012multi} and \citet{bingel2017identifying} approximate multi-task performance by analyzing single-task learning characteristics. An altogether different approach may leverage recent advances in transfer learning focused on understanding task relationships~\citep{taskonomy, achille2019information,dwivedi2019representation,zhuang2020comprehensive, achille2019task2vec}; however, \citet{standley2019tasks} show transfer learning algorithms which determine task similarity do not carry over to the multi-task learning domain and instead propose a multi-task specific framework.

\textbf{Training Dynamics.}  Significant effort has also been invested to improve the training dynamics of MTL systems. In particular, dynamic loss reweighing has achieved performance superior to using fixed loss weights found with extensive hyperparameter search~\citep{uncertainty, guo2018dynamic, liu2019end, gradnorm, moo, pareto_moo}. Another set of methods seeks to mitigate the optimization challenges in multi-task learning by manipulating the task gradients in a number of ways such as (1) modifying the direction of task gradients with the underlying assumption that directional inconsistency of gradients on the shared parameters are detrimental to model convergence and performance~\citep{modulation, suteu2019regularizing}, and (2) altering both the direction and the magnitude of the task gradients~\citep{pcgrad, graddrop, gradvaccine, imtl}. Instead of directly modifying the task gradients during optimization, our work builds upon these approaches by quantifying how a gradient update to the shared parameters would affect training loss and choosing a gradient candidate which most increases positive information transfer. 

\textbf{Looking into the Future.}  Looking at \textit{what could happen} to determine \textit{what should happen} has been used extensively in both the meta-learning~\citep{maml, reptile, chameleon, grant2018recasting, kim2018bayesian} as well as optimization domains~\citep{nesterov27method, hinton1987using, zhang2019lookahead, swa, johnson2013accelerating}. \textit{Lookahead} meta-learning algorithms focusing on validation loss have also been used to improve generalization in multi-task learning systems \citep{wang2020negative}, and our work further adapts this central concept to multi-task learning to both quantify and improve information transfer.

\section{Transference in Multi-Task Learning}
Within the context of a hard-parameter sharing paradigm, tasks collaborate to build a shared feature representation which is then specialized by individual task-specific heads to output a prediction. Accordingly, tasks implicitly transfer information to each other by updating this shared feature representation with successive gradient updates. We can then view transference, or information transfer in multi-task learning, as the effect of a task's gradient update to the shared parameters on another task's loss during training. 

While the shared parameter update using a task's gradient, often but not always, increases the losses of the other tasks in the network, we find the extent to which these losses change to be highly task specific. This indicates certain tasks interact more constructively than others. Further, we notice this effect to be reproducible and nearly unchanged across successive training runs with varying parameter initializations. Motivated by these observations, we derive a quantitative measure of transference, describe how it can be used to determine which tasks should be trained together, and provide empirical analysis of these claims. Later, we build upon these ideas to propose an multi-task learning algorithm that selects among possible gradient candidates to updated the model's shared representation.  

\subsection{Measuring Transference}
\label{sec:measuring_transference}
Consider an $\T$-multitask loss function parameterized by $\{\theta_s\}\cup \{\theta_i\vert\, i \in [\T]\}$ where $\theta_s$ represents the shared parameters and $\theta_i$ represents the task $i\in [\T]$ specific parameters.
Given a batch of examples $\mathcal{X}$, let
\begin{align*}
    L_{\total}(\mathcal{X}, \theta_{s}, \{\theta_{i}\}) = \sum_{i \in [\T]} L_{i}(\mathcal{X}, \theta_{s}, \theta_{i})\, ,
\end{align*}
denote the total loss where $L_i$ represents the non-negative loss of task $i$. For simplicity of notation, we set the loss weight of each task to be equal to 1, though our construction generalizes to arbitrary weightings.

For a given training batch $\mathcal{X}^t$ at time-step $t$, we can first update the task specific parameters $\{\theta^{t+1}_i\}$ using standard gradient updates. We can now define the quantity $\theta_{s\vert i}^{t+1}$ to represent the updated shared parameters after a gradient step with respect to the task $i$. Assuming SGD for simplicity, we have
\begin{align*}
    \theta_{s\vert i}^{t+1} \coloneqq \theta_{s}^{t} - \eta \nabla_{\theta_s}L_i(\mathcal{X}^t, \theta_s^t, \theta_i^t)\, .
\end{align*}
This quantity allows us to calculate a \textit{lookahead} loss using the updated shared parameters while keeping the task-specific parameters as well as the input batch unchanged across different tasks' gradient. That is, in order to assess the effect of the gradient update of task $i$ on a given task $j$, we can compare the loss of task $j$ before and after applying the gradient update on the shared parameters with respect to $i$. In order to eliminate the scale discrepancy among different task losses, we consider the ratio of a task's loss before and after the gradient step on the shared parameters as a scale invariant measure of relative progress. We can then define an asymmetric measure for calculating the \emph{transference} of task $i$ at a given time-step $t$ on task $j$ as 
\begin{align}
    \label{eq:Z-alpha}
    \mathcal{Z}^t_{i \shortarrow j} = 1 - \frac{L_{j}(\mathcal{X}^t, \theta_{s\vert i}^{t+1}, \theta_{j}^{t+1})}{L_{j}(\mathcal{X}^t, \theta_s^t, \theta_j^t)}\, .
\end{align}
Notice that a positive value of $\mathcal{Z}^t_{i \shortarrow j}$ indicates that the update on the shared parameters results in a lower loss on task $j$ than the original parameter values, while a negative value of $\mathcal{Z}^t_{i \shortarrow j}$ indicates that the shared parameter update is antagonistic for this task's performance. Also, note that for $i = j$, our definition of transference encompasses a notion of \emph{self-transference}, i.e. the effect of a task's gradient update on its own loss. This quantity is particularly useful as a baseline to compare against other objectives' transference onto this task. 

\begin{figure}[t!]
    \vspace{-0.1cm}
    \centering
    \begin{center}
    \includegraphics[width=0.36\textwidth]{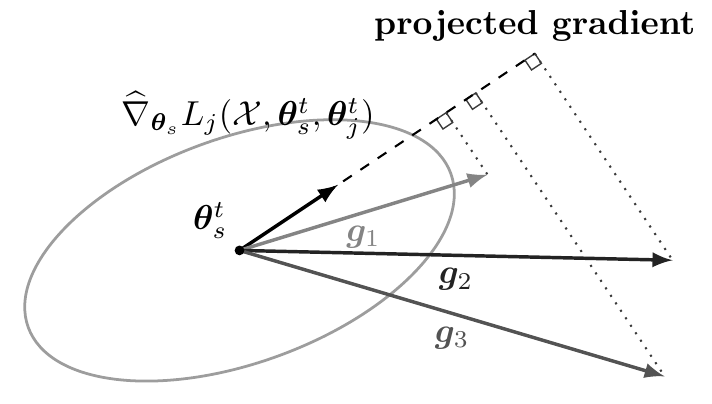}\\
    \end{center}
    \vspace{-0.2cm}
    \caption{Projected gradient along the direction of the task gradient $\hat{\nabla}_{\thet_s}L_j(\mathcal{X}, \thet^t_s, \thet^t_j)$ for three candidate gradients $\bm{g}_1$, $\bm{g}_2$, and $\bm{g}_3$. In this example, $\bm{g}_1$ has the highest alignment, but $\bm{g}_2$ has the largest value of projected gradient. Higher order approximations take into account the second order term which depends on the Hessian of the loss $\nabla^2_{\thet_s}L_j(\mathcal{X}, \thet^t_s, \thet^t_j)$ to penalize orthogonal directions. See the supplementary material for additional detail.}
    \label{fig:proj-gradient}
    \vspace{-0.2cm}
\end{figure}

\subsection{Approximating Transference}
\label{sec:approx}
Computing the transference of task $i$ on objective $j$ requires a backward pass with respect to $\mathcal{L}_{i}$ to compute $\theta_{s|i}^{t+1}$ and then a second forward pass through the network to compute $L_{j}(\mathcal{X}^t, \theta_{s\vert i}^{t+1}, \theta_{j}^{t+1})$. This may become computationally prohibitive for large models, especially as the number of tasks grows. In this section, we derive a simple first order approximation of transference to eliminate the shared parameter update and second forward pass. Ignoring the learning rate $\eta$ for simplicity, a first order Taylor series expansion of transference in~\pref{eq:Z-alpha} yields:
\begin{align*}
     \mathcal{Z}^t_{i \shortarrow j} & = 1 - \frac{L_{j}(\mathcal{X}^t, \theta_{s\vert i}^{t+1}, \theta_{j}^{t+1})}{L_{j}(\mathcal{X}^t, \theta_s^t, \theta_j^t)}\\
    & \approx 1 - \frac{1}{L_{j}(\mathcal{X}^t, \theta_s^t, \theta_j^t)} \Big(L_{j}(\mathcal{X}^t, \theta_s^t, \theta_j^t)\\
    & - \langle \nabla_{\theta_s} L_{j}(\mathcal{X}^t, \theta_s^t, \theta_j^t), \nabla_{\theta_s} L_{i}(\mathcal{X}^t, \theta_s^t, \theta_i^t)\rangle\Big)\\
    & = \frac{\langle \nabla_{\theta_s} L_{j}(\mathcal{X}^t, \theta_s^t, \theta_j^t), \nabla_{\theta_s} L_{i}(\mathcal{X}^t, \theta_s^t, \theta_i^t)\rangle}{L_{j}(\mathcal{X}^t, \theta_s^t, \theta_j^t)}\, ,
\end{align*}
where $\langle\cdot,\cdot\rangle$ denotes inner product. Intuitively, for a fixed task $j$, the transference approximation favors a candidate task $i$ which induces the highest projected gradient value along the direction of $\nabla_{\theta_s} L_{j}(\mathcal{X}^t, \theta_s^t, \theta_j^t)$. This is shown pictorially in \Figref{fig:proj-gradient}. 

\subsection{Task Groupings Based on Transference}
\label{sec:which}
Transference is a measure of the effect of one task's gradient on another objective's loss. We therefore reason that composing task grouping which exhibit strong transference could be an efficient strategy and easy-to-run baseline for determining which tasks should train together in a multi-task learning model. We evaluate this hypothesis on the CelebA~\citep{celeba}, Taskonomy~\citep{taskonomy}, and Meta-World~\citep{metaworld} benchmarks in \Secref{sec:which_exps}.

\section{Increased Transfer MTL}
\begin{figure}
\vspace{-0.4cm}
\begin{center}
\resizebox{0.48\textwidth}{!}{
\begin{minipage}{1\linewidth}
\begin{algorithm}[H]
\begin{algorithmic}[1]
    \STATE Initialize network weights:\,\,  $\{\theta_s\}\cup \{\theta_i\vert\, i \in [\T]\}$
    \STATE Determine the set of gradient candidates:\,\, $\mathcal{J}$
    \FOR{$t = 0, \ldots, T-1$ }
        \STATE Compute per-task loss:\,\;$L_i(\mathcal{X}^t, \theta^t_s, \theta^{t}_i),\,\,  \forall i \in [m]$ \COMMENT{typical forward pass} 
        \STATE Update task-specific parameters:\vspace{-0.2cm}\[\vspace{-0.2cm}\theta_i^{t+1} = \theta_i^t - \eta \nabla_{\theta_i}L_i,\,\, \forall i \in [m]\]\\
        \FOR{$\bm{g}_{\phi} \in \mathcal{J}$}
            \STATE $\theta_{s\vert\phi}^{t+1} = \theta_{s}^{t} - \eta\, \bm{g}_\phi$
            \STATE $\mathcal{Z}^{t}_{\phi} = \sum_{i\in [m]} \Big(1 - \frac{L_{i}(\mathcal{X}^t, \theta_{s\vert\phi}^{t+1}, \theta_{i}^{t+1})}{L_{i}(\mathcal{X}^t, \theta_s^t, \theta_i^{t})}\Big)$
        \ENDFOR
        \STATE Select max transfer gradient mechanism:\vspace{-0.2cm}\[\vspace{-0.3cm}\phi^\star = \argmax_\phi \mathcal{Z}_\phi^t\]
        \STATE Update shared parameters:\vspace{-0.2cm}\[\vspace{-0.5cm}\theta_s^{t+1} = \theta_s^t - \eta\, \bm{g}_{\phi^\star}\]
    \ENDFOR
\end{algorithmic}
\caption{Increased Transfer Multi-Task Learning}
\label{alg:it-mtl}
\end{algorithm}
\end{minipage}
\par
}
\end{center}
\vspace{-0.5cm}
\end{figure}

\label{sec:it-mtl}
A second potential application of transference would be to increase transfer among tasks throughout training. As we can measure the per-batch transference by examining how a gradient update to the shared parameters would increase or decrease the loss of other objectives in the network, we can compare different gradient candidates and apply an update to the shared parameters using the gradient candidate with the highest transference.

Let us define \emph{total transference} at time-step $t$ for gradient candidate $\bm{g}_{\phi}$ induced by the \emph{mechanism} $\phi$ as
\begin{align}
    \label{eq:total}
    \mathcal{Z}^{t}_{\phi} \coloneqq \sum_{j \in [\T]} \mathcal{Z}^t_{\phi \shortarrow j} = \sum_{j \in [\T]} \Big(1 - \frac{L_{j}(X, \theta_{s\vert\phi}^{t+1}, \theta_{j}^{t+1})}{L_{j}(X, \theta_s^t, \theta_{j}^t)}\Big)\, .
\end{align}
In our construction, $\phi$ could be a subset of tasks $[m]$, or a more sophisticated mechanism such as PCGrad~\citep{pcgrad} which, given the state of the model, produces a gradient. The total transference of gradient candidate $\bm{g}_\phi$ can then be viewed as a measure of relative progress across all tasks as a result of applying an update to the shared parameters using this gradient candidate. 
Perhaps surprisingly, $\bm{g}_\phi = \nabla_{\theta_s} L_{\total}(\mathcal{X}, \theta_{s}, \{\theta_{i}\}) =  \sum_{i=1}^m \nabla_{\theta_s} \mathcal{L}_i$ is oftentimes not the gradient candidate which most increases transference. Rather, we find that this particular update can often result in lower transference than the PCGrad gradient~\citep{pcgrad} and even a gradient update with respect to a subset of tasks. 

With this motivation in mind, we present increased transfer multi-task learning ({\color{modelblue} \method}), a parameter-free augmentation to multi-task learning which chooses the gradient that most increases transference.  Specifically, {\color{modelblue} \method} chooses the shared parameter update from a set of gradient candidates which induces the highest total transference in a given minibatch. 

Formally, we define $\mathcal{J}$ as the set of gradient candidates. {\color{modelblue} \method} proceeds by calculating the total transference defined in~\pref{eq:total} for gradient mechanisms $\bm{g}_{\phi} \in \mathcal{J}$ and then applies the gradient update to the shared parameters that induces the highest total transference. Task-specific parameters are updated as normal. The full algorithm is provided in \Algref{alg:it-mtl}. For our empirical results in \Secref{exp:it_mtl} with -PCGrad suffixes, we compose $\mathcal{J} = \{\sum_{i=1}^m \nabla_{\theta_s} \mathcal{L}_i, \bm{g}_{\text{\tiny PCGrad}}\}$; although this set can be augmented with more recent multi-task gradient modification techniques like GradDrop~\citep{graddrop} and/or GradVac~\citep{gradvaccine}. 

\begin{figure*}[t!]
    \vspace{-0.2cm}
    \centering
    \begin{center}
\begin{tabular}{m{0.26\textwidth} m{0.26\textwidth} m{0.26\textwidth}}
 \multicolumn{1}{c}{\small \quad\,\,\,\,\,\,\,\,\,\,\,\, Parameter update in early stage} &  \multicolumn{1}{c}{\small\quad Parameter update in early stage} & \multicolumn{1}{c}{\small\,\,\,\,\,\quad Parameter update in later stage}\\
 \begin{center}
    \subfigure[]{\includegraphics[height=0.95\linewidth]{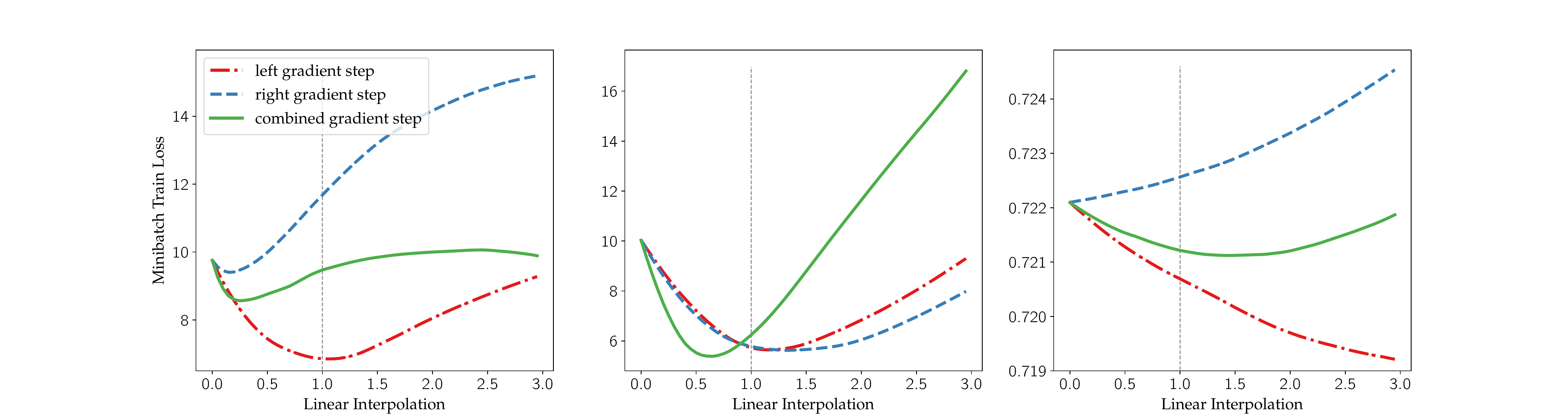}}\end{center} &
    \begin{center}
    \subfigure[]{\includegraphics[height=0.95\linewidth]{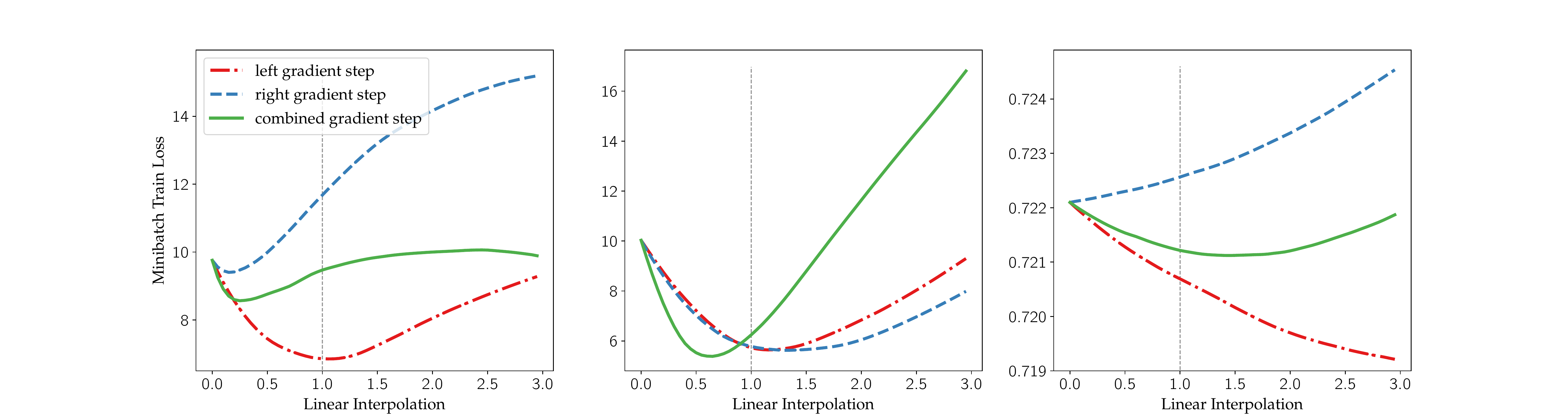}}
    \end{center} 
    & \begin{center}
    \subfigure[]{\includegraphics[height=0.95\linewidth]{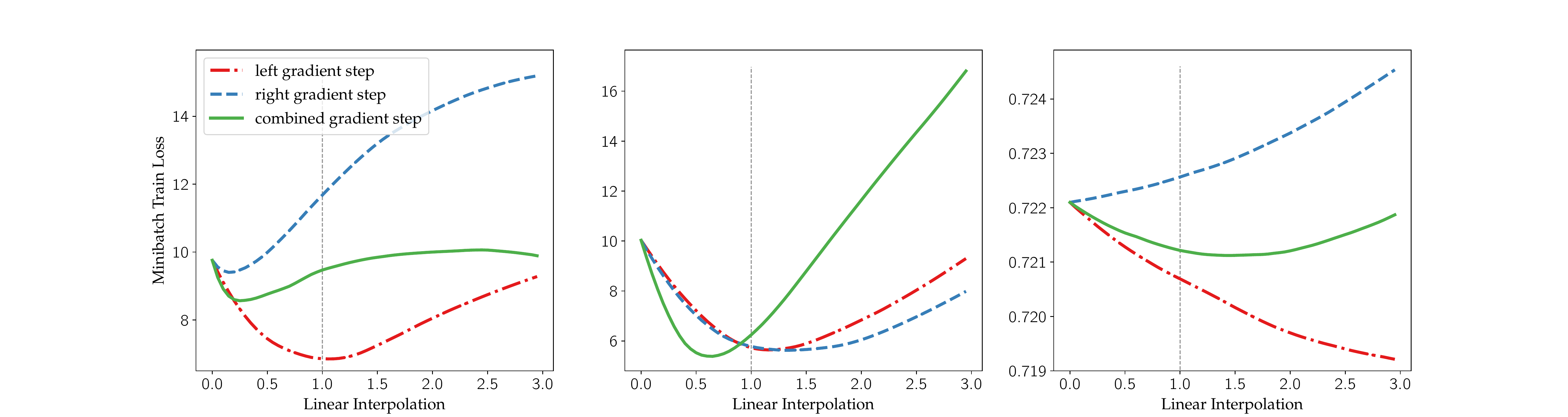}}\end{center}\\
    \end{tabular}
    \end{center}
    \vspace{-0.5cm}
    \caption{Total loss in MultiFashion along a linear interpolation between the current and updated shared parameter values. We extend the interpolation to a magnitude $3\times$ the original update to illustrate the curvature along this direction. The dashed vertical line crosses the loss curves at the original update. In all cases, a step along the left gradient is better.}
    \label{fig:1dlosslandscape}
    \vspace{-0.2cm}
\end{figure*}

\begin{figure*}[t!]

\begin{center}
\begin{tabular}{m{0.28\textwidth} m{0.28\textwidth} m{0.28\textwidth}}
 \multicolumn{1}{c}{\small Left Loss Landscape} & \multicolumn{1}{c}{\small Right Loss Landscape} & \multicolumn{1}{c}{\small Combined Loss Landscape}\\
 \begin{center}
 \subfigure[]{\includegraphics[width=0.25\textwidth, frame]{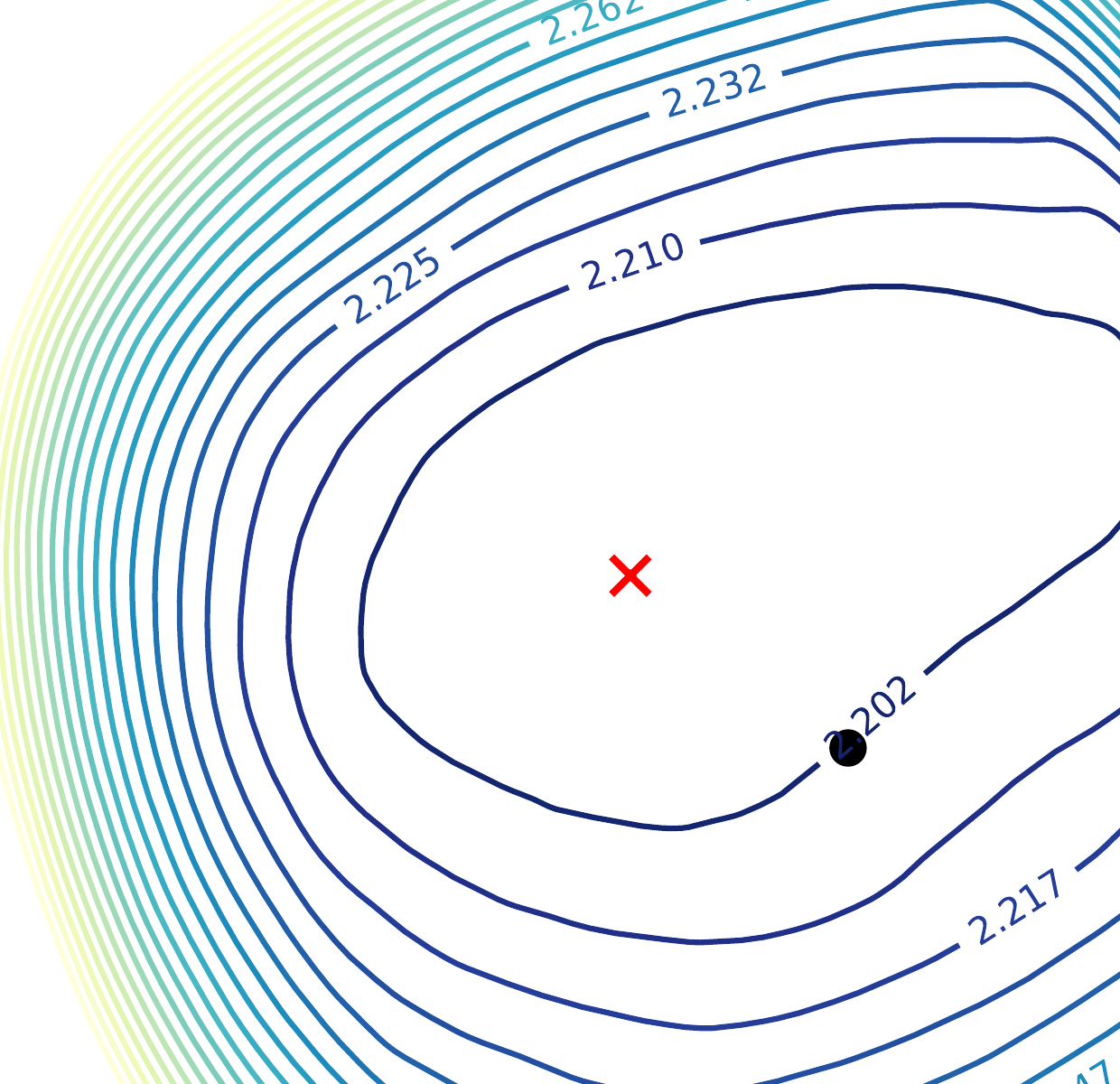}}
 \end{center} & \begin{center} \subfigure[]{\includegraphics[width=0.25\textwidth, frame]{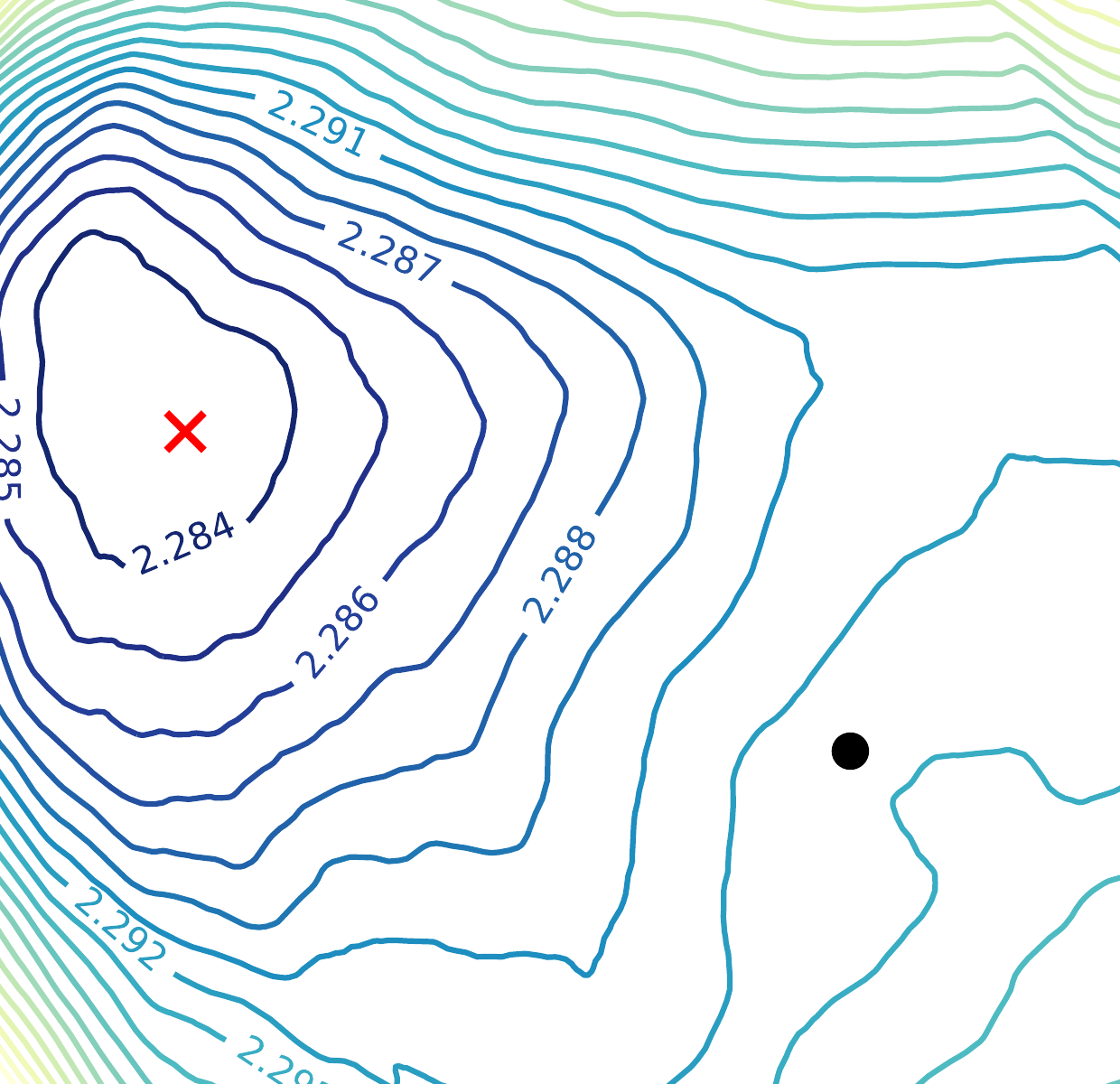}} \end{center} &
 \begin{center}
 \subfigure[]{\includegraphics[width=0.25\textwidth, frame]{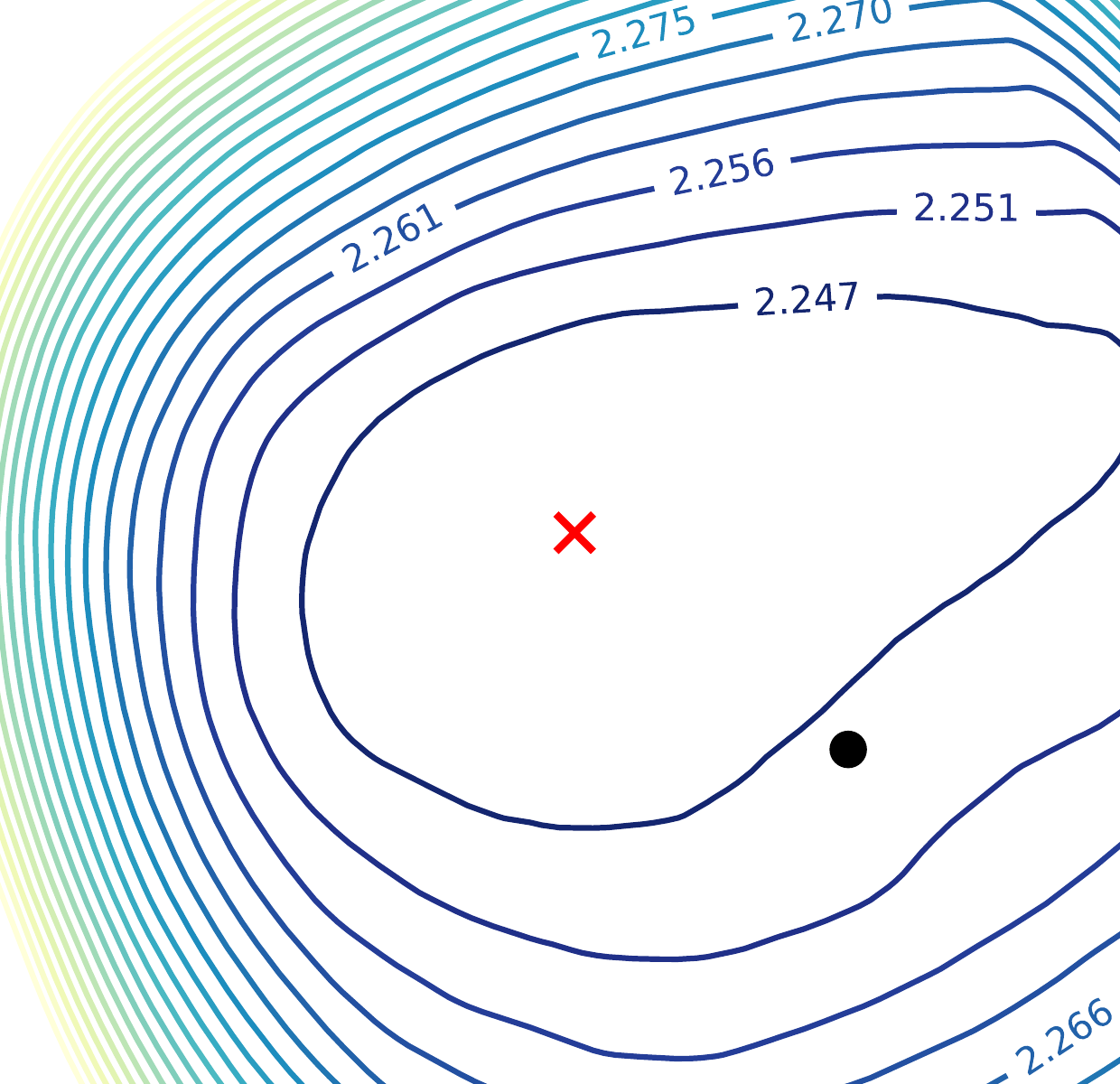}}
 \end{center}
\end{tabular}
\vspace{-0.4cm}
\caption{2-Dimensional loss landscape of the (a)  left digit loss, (b) right digit loss, and (c) the combined (average of left and right) loss in MultiFashion in a step where {\color{modelblue} \method} chooses the left gradient for the shared parameter update. The $\bm{\times}$ symbol indicates the location of the minimum value of the loss in the local landscape and $\bullet$ shows the projected coordinates of the current parameter.}
\label{fig:2dlosslandscape}
\end{center}
\vspace{-0.4cm}
\end{figure*}

\subsection{First Order Approximation}
\label{sec:it_mtl_approx}
Similar to \Secref{sec:approx}, we can reduce computation with a first order Taylor series expansion. The total transference defined in~\pref{eq:total} can be written as
\begin{align*}
    \mathcal{Z}^t_{\phi} & = \sum_{j\in [m]} \mathcal{Z}^t_{\phi \shortarrow j}\\
    & \approx  \langle\sum_{j\in [m]} \frac{ \nabla_{\theta_s} L_{j}(\mathcal{X}^t, \theta_s^t, \theta_j^t)}{L_{j}(\mathcal{X}^t, \theta_s^t, \theta_j^t)}, \bm{g}_\phi\rangle \,,
    \intertext{which can be rewritten as}
    = & \langle \nabla_{\theta_s} \sum_{j\in [m]} \log L_{j}(\mathcal{X}^t, \theta_s^t, \theta_j^t), \bm{g}_\phi\rangle\\
     = & \langle \nabla_{\theta_s} \underbrace{\log \prod_{j\in [m]} L_{j}(\mathcal{X}^t, \theta_s^t, \theta_j^t)}_{\text{log-product loss}}, \bm{g}_\phi\rangle \,.
    \end{align*}
Our {\color{modelblue} \method} approximation computes alignment between the gradients of the candidate tasks with the gradient of the first quantity in the inner product, which we call the ``log-product'' loss. The gradient of the subtasks with the strongest alignment to the gradient of the log-product loss is used to make the final update to the shared parameters.

\subsection{Geometric Analysis}
To explore the mechanism of {\color{modelblue} \method}, we analyze the loss landscape of MultiFashion~\citep{pareto_moo}, a dataset composed by sampling two images from FashionMNIST~\citep{fashion} such that one fashion image is placed in the left while the other is placed in the right with some amount of overlap. \Figref{fig:1dlosslandscape} highlights three cases in which updating the shared parameters with respect to only the left image loss is more effective at decreasing the total loss than updating the shared parameters with respect to both losses. Additional figures and information regarding this analysis can be found in the supplementary material.

\Figref{fig:1dlosslandscape}(a) indicates high curvature in the direction of the right image classification gradient compared to relatively smooth curvature in the direction of the left image classification gradient. \Figref{fig:2dlosslandscape} expands on this analysis in a 2-dimensional contour plot. We hypothesize high curvature in the right image gradient worsens the combined gradient with respect to the total loss. Updating the shared parameters with respect to the left image gradient will result in lower left and right image classification loss than an update using either the right image gradient or combined gradient.

\begin{figure*}[t!]
    \vspace{-0.2cm}
    \centering
    \begin{center}
    \includegraphics[height=0.275\textwidth]{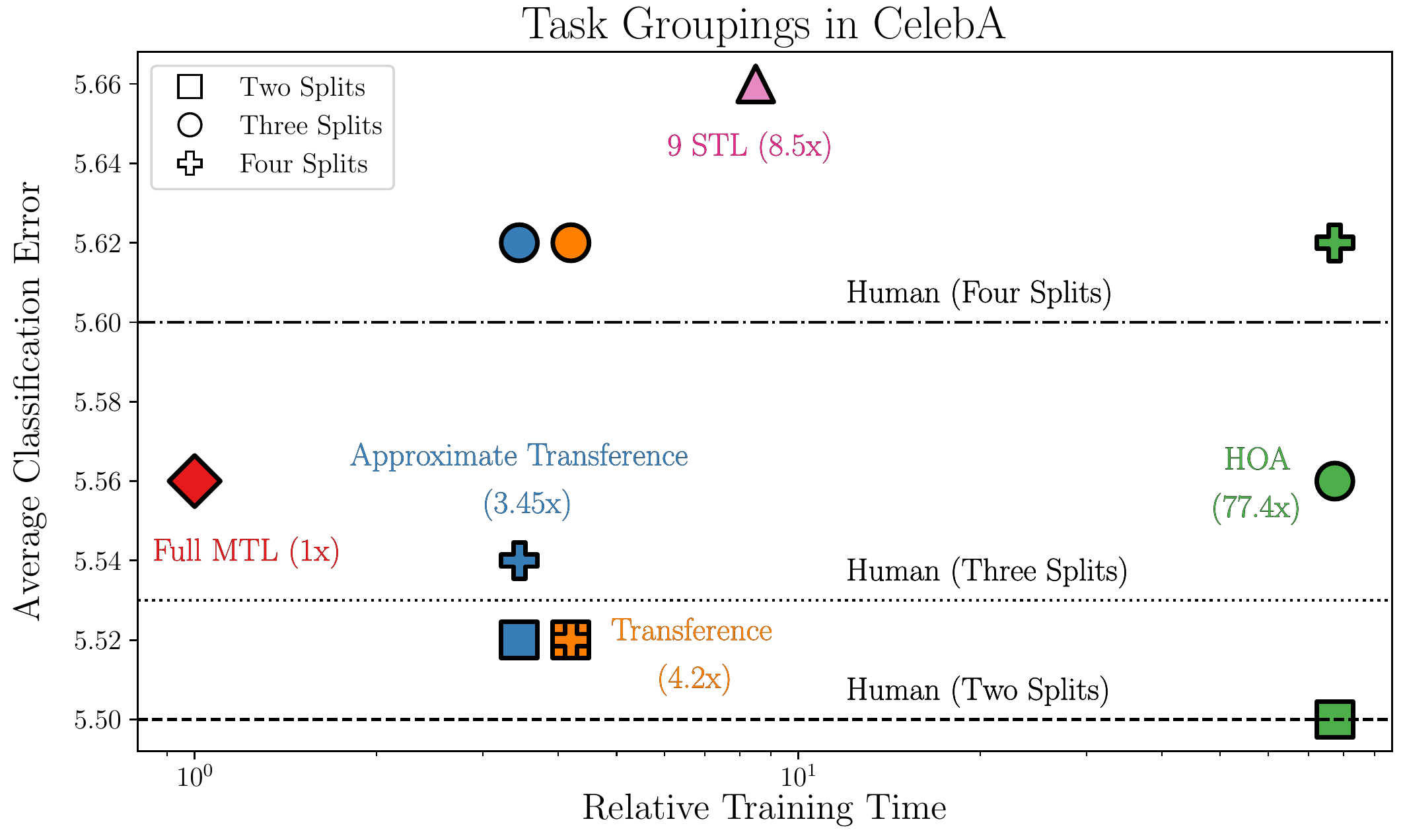}\vspace{0.1cm}
     \includegraphics[height=0.275\textwidth]{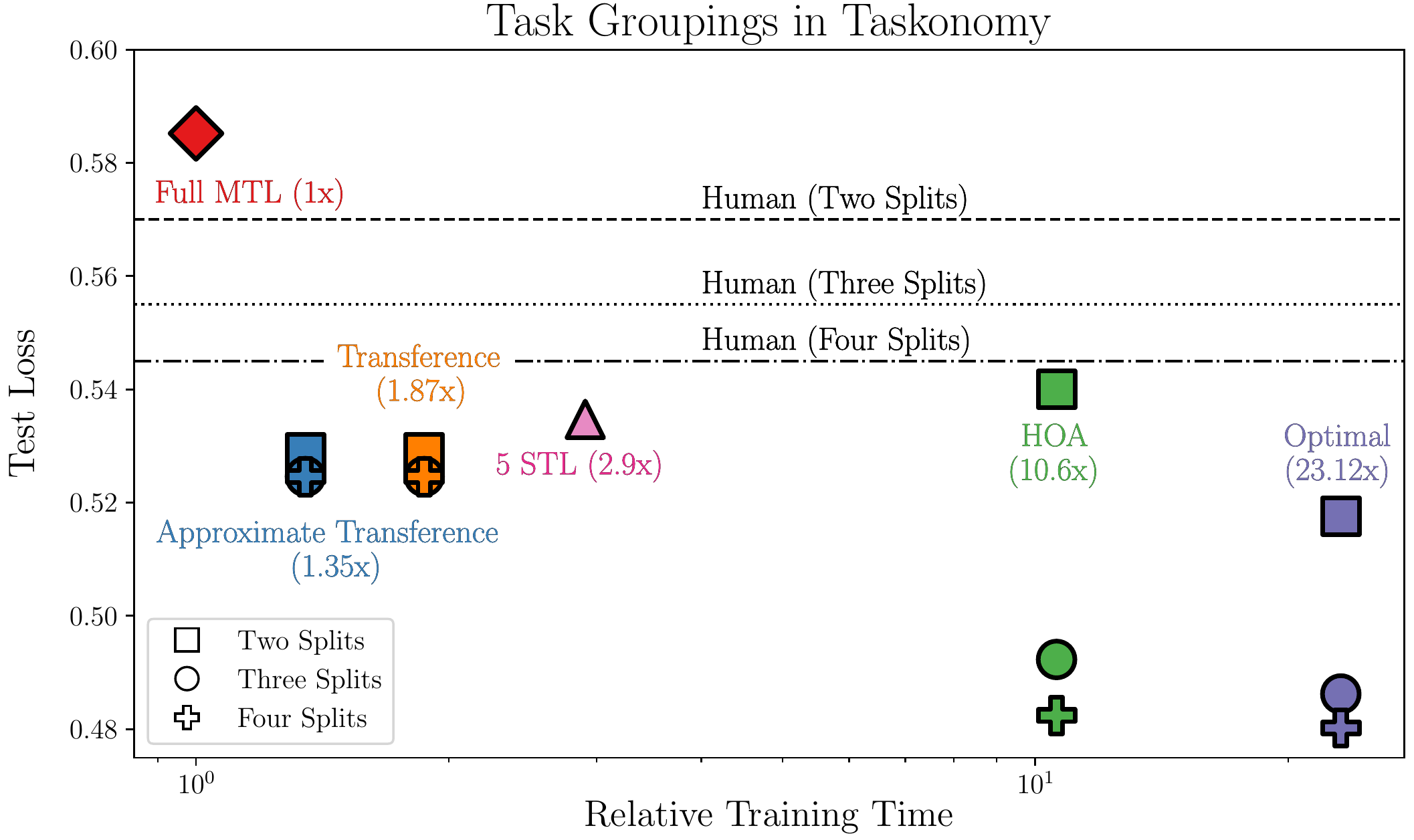}\hfill
    \end{center}
    \vspace{-0.5cm}
    \caption{(Left) average classification error for 2, 3, and 4-split task groupings for the subset of 9 tasks in CelebA. (Right) total test loss for 2, 3, and 4-split task groupings for the subset of 5 tasks in Taskonomy. All models were run on a TeslaV100 instance with the time to train the full MTL model being approximately 83 minutes in CelebA and 73 hours in Taskonomy. Note the x-axis is in log scale.}
    \label{fig:task_groupings}
    \vspace{-0.3cm}
\end{figure*}

In a related instance, \Figref{fig:1dlosslandscape}(b) illustrates an empirically infrequent case when high curvature appears in the direction of the combined gradient, despite relative smoothness along the directions of both right and left image gradients. This result lends weight to framing multi-task learning as multi-objective optimization rather than optimizing a proxy weighted combination of per-task losses since the weighted combination leads to a higher loss than that resultant from applying either the right or left image gradients.

While the first two cases of high curvature occur predominantly during the early stages of training, \Figref{fig:1dlosslandscape}(c) highlights a third scenario which occurs throughout training. In this instance, the curvature of the right image classification loss in the direction of its gradient appears somewhat flat. Further, updating the shared parameters with respect to the right image gradient would sharply increase the left image classification loss. In contrast, the left image gradient is highly effective at minimizing the left image loss as well as the total loss more so than the combined gradient. Interpreting these results, we posit flat regions of curvature in the direction of either objectives' gradient may interfere with the effectiveness of the combined gradient to minimize the total loss.

\vspace{-0.1cm}
\section{Experiments}
\label{sec:expts}
To assess the practical applications of transference, we study its utility in selecting which task should train together as well as the capacity of {\color{modelblue} \method} to improve learning efficiency. We use the CelebA~\citep{celeba}, Taskonomy~\citep{taskonomy}, and Meta-World~\citep{metaworld} benchmarks to evaluate the ability of transference to determine task groupings. Similarly, we evaluate {\color{modelblue} \method} on MultiMNIST, a multitask variant of the MNIST dataset~\citep{mnist}; NYUv2~\citep{nyuv2}; and Movielens~\citep{harper2015movielens}.

\subsection{Task Grouping Experiments}
\label{sec:which_exps}
We analyze the capacity of transference to select task groupings on the CelebA, Taskonomy, and Meta-World datasets. While transference is computed at a per-minibatch level, we can average the transference across minibatches to compute an epoch-level transference score. Integrating across the number of steps in training then provides an overall (scalar) transference value which measures the effect to which one task's gradient decreases (or increases) another task's loss. 

Similar to the method \citet{standley2019tasks} used to select task groupings based on minimizing test loss, we employ a network selection technique to maximize transference given a fixed inference time budget. We define a ``split'' as a single encoder. An inference time budget of two splits supports two shared encoders, each with any number of task-specific decoders. Furthermore, task-specific decoders are not mutually exclusive. One task-specific decoder can help train the shared representation of an encoder while being served from another task grouping. We provide the full algorithm in the supplementary material. All units of time mentioned in subsequent analysis is taken from a cloud instance of Tesla V100s with a beyond sufficient quantity of RAM and CPU. 

\begin{figure*}[t!]
\vspace{-0.2cm}
\begin{center}
\includegraphics[width=0.95\textwidth]{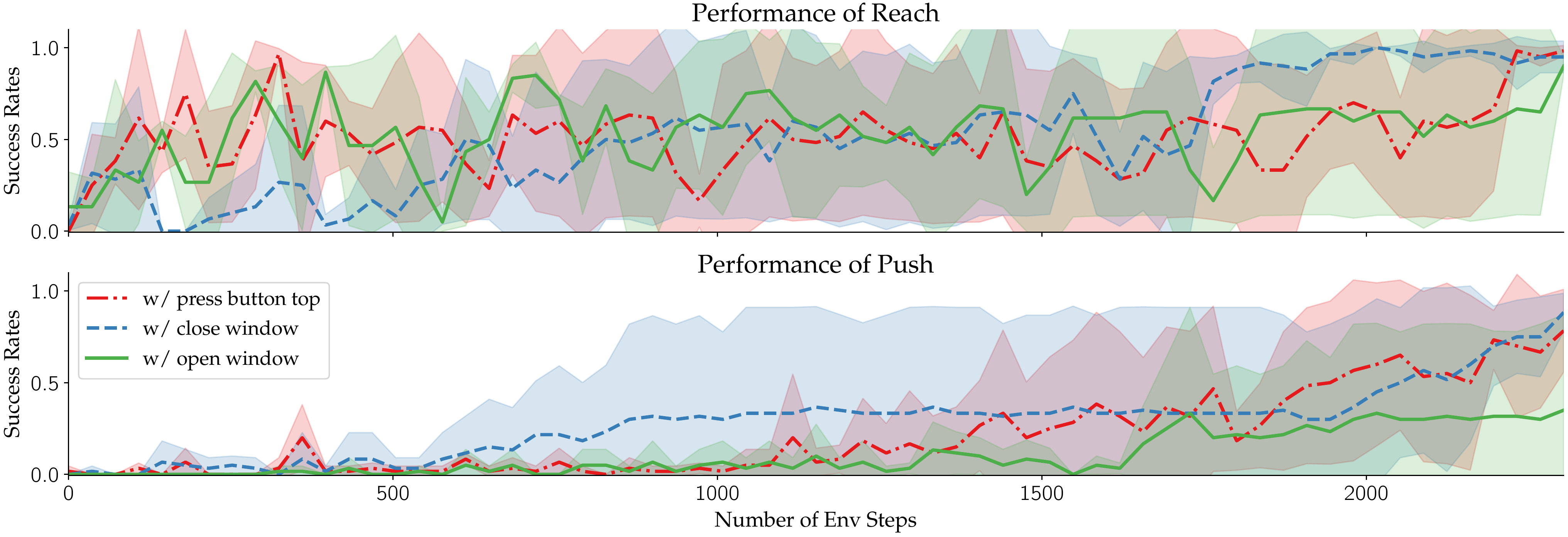}
\end{center}
\vspace{-0.4cm}
\caption{(Top) performance of reach when trained with ``push'' and one of ``press button top'', ``close window'', or ``open window''. (Bottom) performance of ``push'' when trained with reach and one of ``press button top'', ``close window'', or ``open window''. In both cases, the best performance is achieved by co-training with ``close window''. Performance is degraded when co-trained with ``open window''.}
\label{fig:meta}
\vspace{-0.2cm}
\end{figure*}

\subsubsection{CelebA Task Groupings}
We chose a subset of 9 attributes from the 40 possible attributes in CelebA\footnote{The list of chosen attributes is available upon request.}, and evaluate the groupings chosen by transference against a single multitask model, nine single-task models, the HOA baseline proposed by \citep{standley2019tasks}, and a human baseline composed of two industry experts. We report the performance for 2-split, 3-split, and 4-split groupings and include a full description of experimental design in the supplementary material. As this paradigm contains nine tasks, there are 511 (or $2^{|T|} - 1$) possible task combinations and due to computational budget constraints, we were unable to determine optimal groupings which we estimate would take approximately 700 hours of continual training.

Our experimental results are summarized in \Figref{fig:task_groupings}(left). To our surprise, our human experts' excelled in selecting task groupings based on their industry expertise and inherent notions of task similarity when compared with the HOA baseline. Moreover, we find the average performance of the splits formed with transference exceed the performance of the splits found with the HOA approximation while the transference approximation has average performance approximately equal to this baseline. Further, computing transference on CelebA is relatively efficient. It takes approximately 1.24 hours to train a multi-task model, 4.26 hours for the approximation, and 5.2 hours for the complete transference measure on a Tesla V100 GPU. On the other hand, the HOA approximation requires 95.5 hours to complete. This result suggests using transference as a quick to implement and efficient baseline for determining task groupings.

\subsubsection{Taskonomy Task Groupings}
Following the experimental setup of \citet{standley2019tasks}, we evaluate the effectiveness of transference to select task groupings on the ``Segmentic Segmentation'', ``Depth Estimation'', ``Keypoint Detection'', ``Edge Detection'', and ``Surface Normal Prediction'' objectives in Taskonomy.  Our results are summarized in \Figref{fig:task_groupings}(right) and affirm the application of transference as a fast and easy-to-implement baseline to select task groupings. In particular, all transference groupings outperform one multi-task model, five single task models, a human expert baseline, and the HOA 2-split approximation. However, the HOA approximation is significantly more powerful at higher splits, as it can fall back to near-optimal pair groupings rather than rely on possibly imprecise ``higher order'' approximations. Nevertheless, the compute time to determine task groupings with transference is relatively efficient, taking 100 hours for approximate transference and 137 hours for transference when compared with 778 hours of compute for the HOA approximation and 1,696 hours to compute the optimal task groupings.

\subsubsection{Meta-World Task Groupings}
We also consider a multi-task reinforcement learning (RL) problem using the Meta-World benchmark~\citep{metaworld}, which contains 50 qualitatively different robotic manipulation tasks. We select five tasks from Meta-World task suite, namely ``reach'', ``push'', ``press button top'', ``open window'' and ``close window''. We train these five tasks together using the soft actor critic (SAC) \citep{haarnoja2018soft} algorithm with the weights of the critic and the policy shared across all tasks. We compute the transference on the critic loss to produce \Figref{fig:meta_radar} and include additional details on the multi-task RL experiment in the supplementary material.

\begin{figure}[t!]
\vspace{-0.2cm}
\begin{center}
\includegraphics[width=0.47\textwidth]{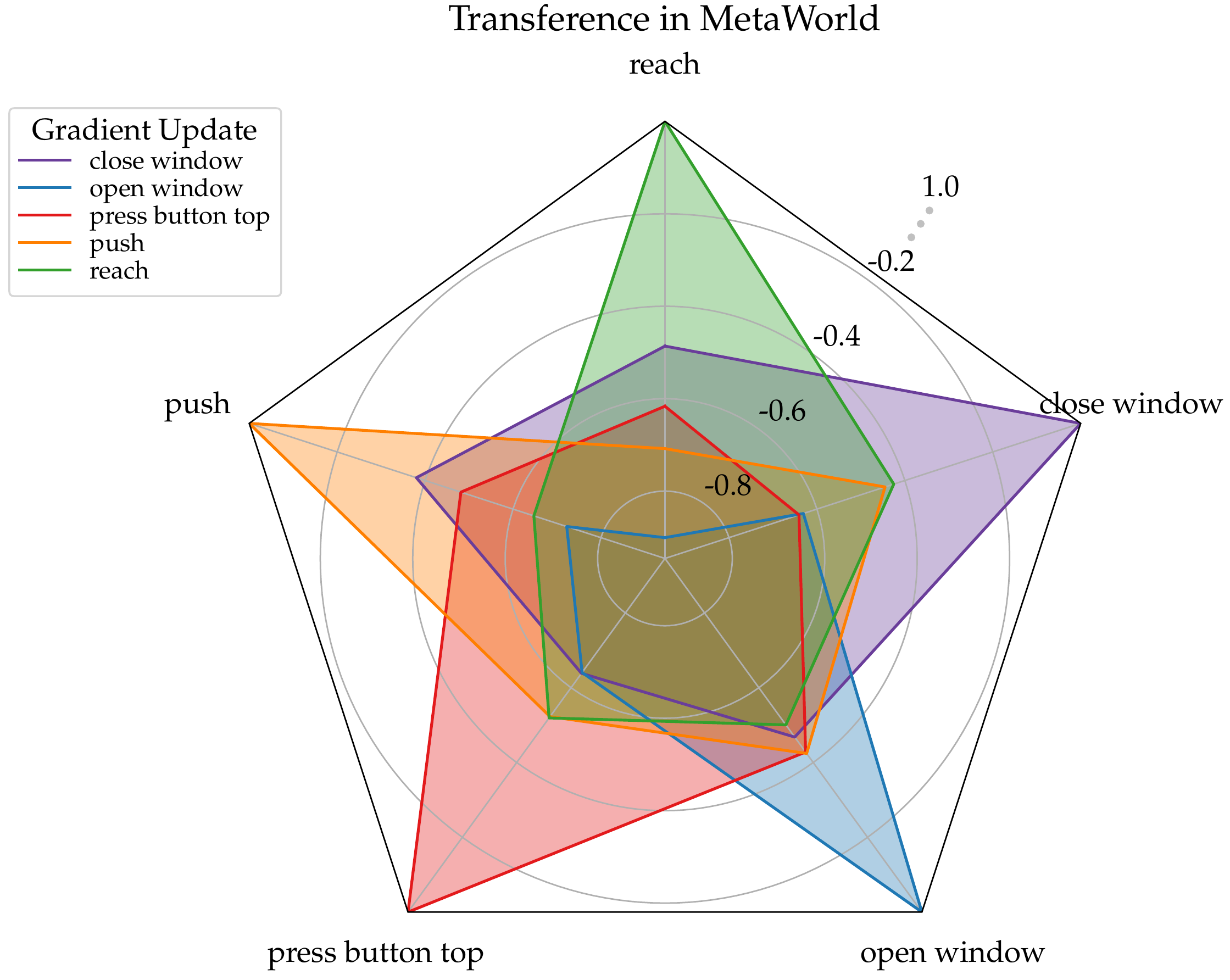}
\end{center}
\vspace{-0.4cm}
\caption{Transference in Meta-World for ``push'', ``reach'', ``press button top'', and ``open window''. For illustration, we normalize the transference along each axis by the task's self-transference. Thus, self-transference becomes 1 for all tasks. Notice ``close window'' manifests high transference on ``push'' and ``reach''.}
\label{fig:meta_radar}
\vspace{-0.6cm}
\end{figure}

\Figref{fig:meta_radar} indicates that ``open window'' exhibits relatively low transference with all tasks while ``close window'' exhibits especially high transference with ``push'' and ``reach''. Accordingly, we group ``push'' and ``reach'' together and then compute the efficacy of these tasks when co-training with ``press button top'', ``open window'', and ``close window''. As shown in \Figref{fig:meta} and as suggested by transference, the success rate of ``reach'' converges more quickly to a significantly higher value when it is co-trained with ``close window'', and marginally faster when it is co-trained with ``press button top'', as compared to co-training with ``open window''. This effect is only magnified when we assess the performance of ``push''. For ``push'', its performance in terms of success rates and data efficiency is greatly increased when co-trained with either ``close window'' or ``press button top'' as compared to co-training with ``open window''. 

\begin{table*}[t!]
\vspace{-0.25cm}
  \centering
  \small
 \caption{13-class semantic segmentation, depth estimation, and surface normal prediction results on the NYUv2 validation dataset averaged across 4 runs. Performance of Cross-Stitch~\citep{cross_stitch} as reported in \citep{liu2019end}. The symbol $\ssymbol{3}$ denotes the approximation described in Section~\ref{sec:it_mtl_approx}, and the best results are highlighted in bold.}
 \vspace{0.1cm}
\resizebox{0.86\textwidth}{!}{%
  \begin{tabularx}{0.81\textwidth}{{l}c*{8}{c}}
  \toprule
  \multicolumn{1}{c}{\multirow{3.5}[4]{*}{Method}} & \multicolumn{2}{c}{Segmentation} & \multicolumn{2}{c}{Depth}  & \multicolumn{5}{c}{Surface Normal}\\
  \cmidrule(lr){2-3} \cmidrule(lr){4-5} \cmidrule(lr){6-10}
  & \multicolumn{2}{c}{\multirow{1.5}[2]{*}{(Higher Better)}} & \multicolumn{2}{c}{\multirow{1.5}[2]{*}{(Lower Better)}}   & \multicolumn{2}{c}{Angle Distance}  & \multicolumn{3}{c}{Within $t^\circ$} \\
  & \multicolumn{1}{c}{} & \multicolumn{1}{c}{}  & \multicolumn{1}{c}{}  & \multicolumn{1}{c}{} & \multicolumn{2}{c}{(Lower Better)} & \multicolumn{3}{c}{(Higher Better)} \\
    &\multicolumn{1}{c}{mIoU}  & \multicolumn{1}{c}{Pix Acc}  & \multicolumn{1}{c}{Abs Err} & \multicolumn{1}{c}{Rel Err} & \multicolumn{1}{c}{Mean}  & \multicolumn{1}{c}{Median}  & \multicolumn{1}{c}{11.25} & \multicolumn{1}{c}{22.5} & \multicolumn{1}{c}{30} \\
\midrule
Cross-Stitch & $14.71$ & $50.23$ & $0.6481$ & $0.2871$ & $33.56$ & $28.58$ & $20.08$ & $40.54$ & $51.97$ \\
MTAN & $22.12$ & $57.16$ & $0.6020$ & $0.2567$ & $31.26$ & $27.27$ & $18.58$ & $41.66$ & $54.89$ \\
PCGrad & $22.63$ & $57.69$ & $0.6143$ & $0.2658$ & $30.85$ & $26.64$ & $19.92$ & $42.88$ & $55.88$ \\
{\color{modelblue} IT-PCGrad} & $\mathbf{22.68}$ & $\mathbf{57.76}$ & $0.6068$  & $0.2600$ & $30.77$  & $26.50$  & $19.86$  & $43.08$ & $56.10$ \\
{\color{modelblue} IT-PCGrad$\ssymbol{3}$}  & $22.21$  & $57.44$  &  $\mathbf{0.5999}$  & $\mathbf{\bf 0.2550}$ & $\mathbf{30.63}$  & $\mathbf{26.11}$ & $\mathbf{20.80}$  & $\mathbf{43.80}$ & $\mathbf{56.65}$ \\
\bottomrule
\end{tabularx}
}
\label{tab:nyuv2}
\vspace{-0.45cm}
\end{table*}

\subsection{IT-MTL Evaluation}
\label{exp:it_mtl}
To assess the effect of increasing total transference throughout training, we evaluate the performance of {\color{modelblue} \method} on the MultiMNIST, NYUv2~\citep{nyuv2}, and MovieLens~\citep{harper2015movielens} datasets. The symbol $\ssymbol{3}$ denotes the approximation described in Section~\ref{sec:it_mtl_approx}.

\textbf{MultiMNIST} is a multi-task version of the MNIST dataset~\citep{mnist} formed by overlapping two digits sampled from the MNIST dataset. To increase comparability, we run our experiments on the same MultiMNIST dataset released by ~\citet{pareto_moo} on a multitask variant of LeNet~\citep{mnist}. Experimental design is detailed in the supplementary material and \citet{pareto_moo} can be referenced for dataset construction. We compare {\color{modelblue} \method} with a single-task baseline, equal weight multi-task learning, PCGrad~\citep{pcgrad}, MGDA-UB~\citep{moo}, and Uncertainty Weighing (UW)~\citep{uncertainty}. 

\Tableref{table:mnist} shows {\color{modelblue} \method} can be used to improve a typical MTL formulation by choosing among the left image, right image, and combined gradient as well as having an additive effect on UW-PCGrad by choosing among the typical multi-task gradient and the PCGrad gradient. Moreover, \Tableref{table:mnist} also highlights the first-order approximation of transference performs similar to that of the non-approximated version. 

\begin{table}[t!]
\centering
\caption{Test Accuracy on MultiMNIST. The test accuracy is averaged over 10 samples. We report standard error, and best results are highlighted in bold. The approximation described in Section~\ref{sec:it_mtl_approx} is denoted with $\ssymbol{3}$.}
\vspace{0.1cm}
\resizebox{\linewidth}{!}{%
\begin{tabular}{l|cc}
\toprule
\multirow{1.5}[3]{*}{Method} &   \multicolumn{2}{c}{MultiMNIST}\\
\cmidrule(lr){2-3} & \multicolumn{1}{c}{Left Image Acc} & \multicolumn{1}{c}{Right Image Acc} \\
\midrule
Single Task Models & $89.40 \pm{0.17}$ & $87.80 \pm{0.22}$ \\
MTL & $89.01 \pm{0.22}$ & $86.11 \pm{0.15}$ \\
PCGrad & $88.92 \pm{0.21}$ & $86.61 \pm{0.22}$\\
{\color{modelblue} \method{}} & $89.12 \pm{0.21}$ & $86.35 \pm{0.23}$ \\
MGDA-UB & $89.38 \pm{0.16}$ & $86.81 \pm{0.46}$ \\ 
UW-MTL & $90.77 \pm{0.13}$ & $88.36 \pm{0.12}$ \\ 
UW-PCGrad & $90.77 \pm{0.11}$ & $88.36 \pm{0.09}$ \\
{\color{modelblue} IT-UW-MTL} & $90.71 \pm{0.18}$ & $\mathbf{88.51 \pm{0.13}}$ \\
{\color{modelblue} IT-UW-PCGrad} & $\mathbf{90.94 \pm{0.11}}$ & $\mathbf{88.61 \pm{0.14}}$ \\
{\color{modelblue} IT-UW-PCGrad$\ssymbol{3}$} & $\mathbf{90.92 \pm{0.09}}$ & $\mathbf{88.60 \pm{0.14}}$ \\
\bottomrule
\end{tabular}
}
\label{table:mnist}
\vspace{-0.4cm}
\end{table}

\textbf{NYUv2} is a popular large-scale computer vision dataset. It is composed of RGB-D images of indoor scenes and supports the modeling of 13-class semantic segmentation, true depth estimation, and surface normal prediction. We follow the procedure of \citep{liu2019end} and directly utilize their Multi-Task Attention Network (MTAN) architecture to evaluate the performance of {\color{modelblue} \method}. For both {\color{modelblue} IT-PCGrad} and the first-order approximation {\color{modelblue} IT-PCGrad$\ssymbol{3}$}, $\mathcal{J}$ is composed of the typical MTL gradient and the PCGrad gradient.

\Tableref{tab:nyuv2} shows the performance of PCGrad surpssses the performance of MTL in Segmentation and Normals while lagging behind in Depth. However, using {\color{modelblue} \method} to dynamically select between the PCGrad gradient and the MTL gradient throughout training improves nearly all measures of performance when compared with the PCGrad baseline. Similarly, {\color{modelblue} \method$\ssymbol{3}$} surpasses both PCGrad and MTL baselines in Depth and Surface Normals while marginally decreasing the performance of Segmentation. 

\begin{table}[t!]
\centering
\caption{Test Accuracy (\%) on MovieLens averaged over 5 runs. We report standard error, and best results are highlighted in bold. $\ssymbol{3}$ denotes the approximation described in Section~\ref{sec:it_mtl_approx}.}
\vspace{0.1cm}
\resizebox{0.9\linewidth}{!}{%
\label{tab:movielens}
\begin{tabular}{l|cc}
\toprule
Method & Watch Acc. & Rating Acc. \\
\midrule
MTL & $80.97 \pm{0.14}$ & $\mathbf{73.99 \pm{0.02}}$ \\
UW-MTL & $81.84 \pm{0.08}$ & $73.95 \pm{0.07}$ \\
UW-PCGrad & $81.93 \pm{0.04}$ & $73.93 \pm{0.02}$ \\
{\color{modelblue} UW-IT-PCGrad$\ssymbol{3}$} & $\mathbf{82.15 \pm{0.11}}$ & $\mathbf{73.97 \pm{0.05}}$ \\
\bottomrule
\end{tabular}
}
\label{table:movielens}
\vspace{-0.4cm}
\end{table}

\textbf{The MovieLens dataset} contains 1 million movie ratings from 6000 users on 4000 movies. Following the experimental design of \citep{wang2020small} and to mimic recommendation systems, we construct user-movie pairs to predict if the user watched the associated movie, and if so, the rating given to the movie by this user. The first task is formulated as a binary classification while the latter is formulated as a regression. \Tableref{tab:movielens} again shows that combining Uncertainty Weights with PCGrad and {\color{modelblue} \method} yields improved performance on both tasks over the the UW-MTL or UW-PCGrad baselines.

\vspace{-0.1cm}
\section{Conclusion}
In this work, we take a first step towards quantifying information transfer in multi-task learning. We develop a measure to quantify transference and leverage this quantity to determine which tasks should be trained together as well as develop a method which improves multi-task learning efficiency and performance. Future research on transference can incorporate this measure into a continuous-space learning algorithm, or guide the development of flexible architectures to improve multi-task learning performance. 

\newpage

\bibliography{references}
\bibliographystyle{icml2021}

\clearpage

\twocolumn[   
\icmltitle{Measuring and Harnessing Transference in Multi-Task Learning \\ (Supplementary Material)}
]

\appendix

\section{Second Order Approximation of Transference}
In this section, we consider a second order approximation of transference in~\pref{eq:Z-alpha}. For simplicity, we write $\bm{g}_i \coloneqq \nabla_{\theta_s} L_{i}(\mathcal{X}^t, \theta_s^t, \theta_i^t)$. This yields
\begin{align*}
L_{j}(\mathcal{X}^t, \theta_{s\vert i}^{t+1}, \theta_{j}^{t+1}) & \approx L_{j}(\mathcal{X}^t, \theta_s^t, \theta_j^t)\\
& - \langle \nabla_{\theta_s} L_{j}(\mathcal{X}^t, \theta_s^t, \theta_j^t),\, \bm{g}_i\rangle\\
& + \sfrac{1}{2}\, \bm{g}^\top_i\,  \nabla^2_{\theta_s}L_{j}(\mathcal{X}^t, \theta_s^t, \theta_j^t)\, \bm{g}_i\, ,
\end{align*}
in which, $\nabla^2_{\theta_s}L_{j}(\mathcal{X}^t, \theta_s^t, \theta_j^t)$ is the Hessian of the loss $L_j$. Plugging back into the definition, we have
\begin{align*}
    & \mathcal{Z}^t_{i \shortarrow j} = 1 - \frac{L_{j}(\mathcal{X}^t, \theta_{s\vert i}^{t+1}, \theta_{j}^{t+1})}{L_{j}(\mathcal{X}^t, \theta_s^t, \theta_j^t)}\\
    & \approx \frac{\langle \nabla_{\theta_s} L_{j}(\mathcal{X}^t, \theta_s^t, \theta_j^t),\, \bm{g}_i\rangle - \sfrac{1}{2}\, \bm{g}^\top_i\,  \nabla^2_{\theta_s}L_{j}(\mathcal{X}^t, \theta_s^t, \theta_j^t)\, \bm{g}_i}{L_{j}(\mathcal{X}^t, \theta_s^t, \theta_j^t)}.
\end{align*}
Intuitively for a fixed task $j$, the second approximation of transference favors a direction $\bm{g}_i$ which has the largest projected gradient along the direction of the task gradient $\nabla_{\theta_s} L_{j}(\mathcal{X}^t, \theta_s^t, \theta_j^t)$. However, it penalizes directions that are orthogonal to $\nabla_{\theta_s} L_{j}(\mathcal{X}^t, \theta_s^t, \theta_j^t)$ and thus have no effect on increasing the projection. The amount of penalty is proportional to the eigendirections of the Hessian matrix $\nabla^2_{\theta_s}L_{j}(\mathcal{X}^t, \theta_s^t, \theta_j^t)$. In our first order approximation of transference used in the experiments, we omit the second term and pick the task with the highest projected gradient in each iteration.

\section{Network Selection Algorithm}
We follow the network selection technique detailed in \citet{standley2019tasks} to find the optimal task groupings given a fixed inference-time computational budget of splits and the performance (or an approximation of the performance) of every combination of tasks. This problem is NP-hard (reduction from Set-Cover), but can be solved efficiently with a branch-and-bound-like algorithm as detailed in \citet{standley2019tasks} or with a binary integer programming solver as done in \citet{taskonomy}. 

Following the procedure outlined in \Secref{sec:which_exps} to convert mini-batch transference to an overall training-time score, we define $t(b,a)$ to be the transference score from task $b$ onto task $a$. In other words, $t(b,a)$ measures the effect of applying the gradient with respect to the loss of task $b$ on the shared parameters, and measuring how this update would affect the loss of task $a$ throughout the course of training.

For the purposes of using transference in a network selection algorithm, we define $$\hat{t}(b,a) = 1.0 - \frac{t(b,a)}{t(a,a)}$$ and minimize this quantity which is equivalent to maximizing each task's normalized transference score. We then compute a group's transference by summing across normalized transference pairs. For example, in a group consisting of tasks $\{a, b, c\}$, we define the total transference of this group as
$$\frac{\hat{t}(b,a) + \hat{t}(c,a)}{2} + \frac{\hat{t}(a,b) + \hat{t}(c,b)}{2} + \frac{\hat{t}(a,c) + \hat{t}(b,c)}{2}$$

We can then use the branch-and-bound algorithm in ~\citet{standley2019tasks} or even a brute force solution to determine the combinations of groups which would result in the highest normalized transference given a fixed inference-time budget.

\begin{table}[t!]
\vspace{-0.5cm}
  \centering
  \small

 \caption{7-class semantic segmentation and depth estimation results on CityScapes validation dataset averaged across 6 runs. The symbol {\color{modelblue} $\ssymbol{3}$} denotes the approximation described in Section~\ref{sec:it_mtl_approx}, and the best results are highlighted in bold.}
 \vspace{0.1cm}
  \resizebox{\linewidth}{!}{%
  \begin{tabularx}{\linewidth}{c*{4}{c}}
  \toprule
  \multicolumn{1}{c}{\multirow{3.5}[4]{*}{Method}} & \multicolumn{2}{c}{Segmentation} & \multicolumn{2}{c}{Depth}\\
  \cmidrule(lr){2-3} \cmidrule(lr){4-5}
  & \multicolumn{2}{c}{\multirow{1.5}[2]{*}{(Higher Better)}} & \multicolumn{2}{c}{\multirow{1.5}[2]{*}{(Lower Better)}} \\
  & \multicolumn{1}{c}{} & \multicolumn{1}{c}{}  & \multicolumn{1}{c}{}  & \multicolumn{1}{c}{} \\
    &\multicolumn{1}{c}{mIoU}  & \multicolumn{1}{c}{Pix Acc}  & \multicolumn{1}{c}{Abs Err} & \multicolumn{1}{c}{Rel Err}\\
\midrule
UW-MTAN & $59.84$ & $91.67$ & $\mathbf{0.0163}$ & $0.3126$\\
UW-PCGrad & $59.83$ & $91.65$ & $0.0171$ & $0.3166$\\
{\color{modelblue} UW-IT-PCGrad$\ssymbol{3}$} & $\mathbf{59.87}$ & $\mathbf{91.68}$ & $0.0168$ & $\mathbf{0.3064}$\\
\bottomrule
\end{tabularx}
}
\label{tab:cityscapes}
\vspace{-0.3cm}
\end{table}

\section{Additional Experiments}
\textbf{Cityscapes} is a large-scale computer vision dataset composed of high resolution street-view images. It supports 7-class semantic segmentation and ground-truth inverse depth estimation. We follow the procedure of \citet{liu2019end} and directly utilize their Multi-Task Attention Network (MTAN) architecture to evaluate the performance of {\color{modelblue} \method}. For {\color{modelblue} IT-PCGrad$\ssymbol{3}$}, $\mathcal{J}$ is composed of the typical MTL gradient and the PCGrad gradient.

\Tableref{tab:cityscapes} indicates that PCGrad by itself will not improve performance over the UW-MTL baseline; however, using transference to select between the typical MTL gradient and the PCGrad gradient slightly improves Segmentation performance over the UW-MTL and UW-PCGrad baselines and significantly improves the Relative Error measurement of Depth performance. While {\color{modelblue} \method} is limited by choosing among a set of gradient candidates, the results in \Tableref{tab:cityscapes} suggest {\color{modelblue} \method} can improve performance even when the candidate gradient(s) by themselves may be detrimental to model performance when compared with a MTL baseline.

\textbf{MultiFashion} is a multi-task version of the FashionMNIST dataset~\citep{fashion} formed by overlapping two fashion images sampled from the FashionMNIST datatset. To increase comparability, we run our experiments on the same MultiFashion dataset released by~\citet{pareto_moo} on a multitask variant of LeNet~\citep{mnist}. 

\Tableref{table:fashion} shows choosing among the left image gradient, the right image gradient, and the combined gradient greatly improves the performance of MTL. Similarly, {\color{modelblue} IT-UW-MTL} improves both left and right image accuracy over UW-MTL, and {\color{modelblue} IT-UW-PCGrad} also improves both left and right image accuracy over UW-PCGrad. However, we found very high inter-run variance, which we believe is due to high curvature in the loss landscape. Likewise, we find the transference approximation is ineffective at replicating the performance of transference as it relies on a first-order Taylor series to approximation the \emph{lookahead loss} which is inaccurate in regions of high curvature.

\begin{table}[t!]
\vspace{-0.3cm}
\centering
\caption{Test Accuracy on MultiFashion. The test accuracy is averaged over 10 samples. We report standard error, and best results are highlighted in bold. The approximation described in Section~\ref{sec:it_mtl_approx} is denoted with $\ssymbol{3}$.}
\vspace{0.3cm}
\resizebox{\linewidth}{!}{%
\begin{tabular}{l|cc}
\toprule
\multirow{1.5}[3]{*}{Method} &   \multicolumn{2}{c}{MultiFashion}\\ 
\cmidrule(lr){2-3} & \multicolumn{1}{c}{Left Image Acc} & \multicolumn{1}{c}{Right Image Acc} \\
\midrule
Single Task Models & $80.31 \pm{0.33}$ & $78.63 \pm{0.64}$ \\
MTL & $78.09 \pm{0.56}$ & $77.15 \pm{0.63}$ \\
PCGrad & $77.19 \pm{0.84}$ & $75.42 \pm{1.03}$ \\
{\color{modelblue} \method{}} & $78.89 \pm{0.37}$ & $78.22 \pm{0.28}$ \\
MGDA-UB & $78.59 \pm{0.35}$ & $77.08 \pm{0.54}$ \\ 
UW-MTL & $80.82 \pm{0.13}$ & $79.95 \pm{0.23}$ \\ 
UW-PCGrad & $80.87 \pm{0.18}$ & $80.33 \pm{0.16}$ \\
{\color{modelblue} IT-UW-MTL} & $80.83 \pm{0.22}$ & $80.36 \pm{0.21}$ \\
{\color{modelblue} IT-UW-PCGrad} & $\mathbf{81.10 \pm{0.17}}$ & $\mathbf{80.65 \pm{0.26}}$ \\
{\color{modelblue} IT-UW-PCGrad$\ssymbol{3}$} & $\mathbf{80.98 \pm{0.19}}$ & $80.27 \pm{0.15}$ \\
\bottomrule
\end{tabular}
}
\vspace{-0.4cm}
\label{table:fashion}
\end{table}

\section{Experimental Design}
In this section, we provide additional details into the design of our experiments with the goal of facilitating reproducibility. We will later open source our code to facilitate reproducibility and increase accessibility to MTL research.

\subsection{CelebA}
We accessed the CelebA dataset publicly available on TensorFlow datasets {\color{urlorange} \small \url{https://www.tensorflow.org/datasets/catalog/celeb_a}} and filtered the 40 annotated attributes down to a set of 9 attributes for our analysis. A list of the chosen attributes used in our analysis are available upon request. Our experiments were run on a combination of Keras~\citep{chollet2018keras} and TensorFlow~\citep{abadi2016tensorflow} running in a Tesla V100 backed Jupyter Notebook.

The encoder architecture is based loosely on ResNet 18~\citep{resnet} with task-specific decoders being composed of a single projection layer. A coarse architecture search revealed adding additional layers to the encoder and decoder did not meaningfully improve model performance. A learning rate of 0.0005 is used for 100 epochs, with the learning rate being halved every 15 epochs. The learning rate was tuned on the validation split of the CelebA dataset over the set of $\{0.00005, 0.0001, 0.0005, 0.001, 0.005, 0.01\}$. We train until the validation increases for 10 consecutive epochs, load the parameters from the best validation checkpoint, and evaluate on the test set. We use the splits default to TensorFlow datasets of (162,770, 19,867, 19,962) for (Train, Valid, Test).

All measurements listed in the experiments are averaged across three independent runs; although we find inter-run variance to be very low.

\begin{table*}[t!]
\centering
\resizebox{0.85\linewidth}{!}{%
\begin{tabular}{l|c|c|c|c|c|c}
\toprule
Method & Budget & Group 1 & Group 2 & Group 3 & Group 4 & Avg. Accuracy \\ 
\midrule
\multirow{3}{*}{HOA} & 2-split & \{t1, t5, t6, t7, t8\} & \{t2, t3, t4, t9\} & --- & --- & 94.50 \\
& 3-split & \{t1, t5, t6, t7, t8\} & \{t2, t9\} & \{t3, t4\} & --- & 94.44 \\
& 4-split & \{t3, t4, t6, t8\} & \{t1, t7, \textbf{t8}\} & \{t2, t9\} & \{t5, \textbf{t7}\} & 94.38 \\
\hline
\multirow{3}{*}{Transference} & 2-split & \{t1, t2, t3, t4, t7, t8, t9\} & \{t5, t6, t7\} & --- & --- & 94.48 \\
& 3-split & \{t2, t3, t4, t8, t9\} & \{t1, \textbf{t7}\} & \{t5, t6, t7\} & --- & 94.38 \\
& 4-split & \{t2, t3, t4, t7, t8, t9\} & \{t5, t6, t7\} & \{t1, \textbf{t7}\} & \{\textbf{t2}, \textbf{t4}\} & 94.48 \\
\hline
\multirow{3}{*}{Transference$\ssymbol{3}$} & 2-split & \{t1, t2, t3, t4, t7, t8, t9\} & \{t5, t6, t7\} & --- & --- & 94.48 \\
& 3-split & \{t2, t3, t4, t8, t9\} & \{t1, t7\} & \{t5, t6\} & --- & 94.38 \\
& 4-split & \{t2, t3, t4, t8, t9\} & \{t1, t7\} & \{t5, t6\} &  \{\textbf{t2}, \textbf{t4}\} & 94.46 \\
\hline
\multirow{3}{*}{Human Expert 1} & 2-split & \{t1, t5, t6, t7, t8\} & \{t2, t3, t4, t9\} & --- & --- & 94.50 \\
& 3-split & \{t2, t3, t4, t9\} & \{t1, t7, t8\} & \{t5, t6\} & --- & 94.47 \\
& 4-split & \{t1, t7, t8\} & \{t2, t3, t4\} & \{t5, t6\} &  \{t9\} & 94.40 \\
\hline
\multirow{3}{*}{Human Expert 2} & 2-split & \{t1, t5, t6, t7, t8\} & \{t2, t3, t4, t9\} & --- & --- & 94.50 \\
& 3-split & \{t2, t3, t4, t9\} & \{t1, t5, t6, t7\} & \{t8\} & --- & 94.48 \\
& 4-split & \{t1, t5, t6, t7 \}& \{t2, t3, t4, t8\} & \{t8\} & \{t9\} & 94.42 \\
\bottomrule
\end{tabular}
}
\vspace{-0.2cm}
\caption{Chosen task groupings in CelebA. A task within a group is highlighted in bold when this task appears in multiple groups, and it is chosen to ``serve'' from this assigned task grouping. Human experts 1 and 2 are industry professionals with extensive familiarity with large multi-task networks.}
\label{table:celeba_groupings}
\end{table*}

\subsection{Taskonomy}
Our experiments mirror the settings and hyperparameters of ``Setting 3'' in \citet{standley2019tasks} by directly implementing transference and its approximation in the framework provided by the author's official code release ({\color{urlorange} \small \url{https://github.com/tstandley/taskgrouping}} at hash dc6c89c269021597d222860406fa0fb81b02a231). However to reduce computational requirements and increase accessibility, we replace the approximately 2.4 TB full+ Taskonomy split used by \citet{standley2019tasks} with an augmented version of the medium Taskonomy dataset split by filtering out buildings with corrupted images and adding additional buildings to replace the corrupted ones. The list of buildings used in our analysis is encapsulated within our released code.

\begin{table*}[t!]
\centering
\resizebox{0.7\linewidth}{!}{%
\begin{tabular}{l|c|c|c|c|c|c}
\toprule
Method & Budget & Group 1 & Group 2 & Group 3 & Group 4 & Total Test Loss \\ 
\midrule
\multirow{3}{*}{Optimal} & 2-split & \{s, d, n\} & \{n, k, t\} & --- & --- & 0.5176 \\
& 3-split & \{s, d, n\} & \{s, \textbf{n}, t\} & \{n, k\} & --- & 0.4862 \\
& 4-split & \{s, d, n\} & \{\textbf{s}, \textbf{n}\} & \{n, \textbf{k}\} & \{t, k\} & 0.4802 \\
\hline
\multirow{3}{*}{HOA} & 2-split & \{s, d, \textbf{n}, t\} & \{n, k\} & --- & --- & 0.5400 \\
& 3-split & \{s, d, \textbf{n}\} & \{n, \textbf{k}\} & \{t, k\} & --- & 0.4923 \\
& 4-split & \{s, n\} & \{d, \textbf{n}\} & \{n, \textbf{k}\} & \{t, k\} & 0.4824 \\
\hline
\multirow{3}{*}{Transference} & 2-split & \{s, d, n\} & \{t, k\} & --- & --- & 0.5288 \\
& 3-split & \{\textbf{s}, d\} & \{\textbf{d}, n\} & \{t, k\} & --- & 0.5246 \\
& 4-split & \{\textbf{s}, d\} & \{s, n\} & \{\textbf{d}, \textbf{n}\} & \{t, k\} & 0.5246 \\
\hline
\multirow{3}{*}{Transference$\ssymbol{3}$} & 2-split & \{s, d, n\} & \{t, k\} & --- & --- & 0.5288 \\
& 3-split & \{\textbf{s}, d\} & \{\textbf{d}, n\} & \{t, k\} & --- & 0.5246 \\
& 4-split & \{\textbf{s}, d\} & \{s, n\} & \{\textbf{d}, \textbf{n}\} & \{t, k\} & 0.5246 \\
\hline
\multirow{3}{*}{Human Expert 1} & 2-split & \{\textbf{s}, d, n\} & \{s, t, k\} & --- & --- & 0.5483 \\
& 3-split & \{\textbf{s}, \textbf{d}, n\} & \{s, t, \textbf{k}\} & \{s, d, k\} & --- & 0.5483 \\
& 4-split & \{\textbf{s}, n, \textbf{t}\} & \{d, n, k\} & \{\textbf{d}, \textbf{n}\} &  \{s, t\} & 0.5362 \\
\hline
\multirow{3}{*}{Human Expert 2} & 2-split & \{s, d, n, t\} & \{\textbf{s}, \textbf{d}, k\} & --- & --- & 0.5871 \\
& 3-split & \{\textbf{s}, \textbf{n}, \textbf{t}\} & \{s, \textbf{d}, k\} & \{d, n, t\} & --- & 0.5575 \\
& 4-split & \{s, d, n, t, \textbf{k}\} & \{\textbf{s}, \textbf{n}, \textbf{t}\} & \{d, n, t\} &  \{s, \textbf{d}, k\} & 0.5453 \\
\bottomrule
\end{tabular}
}
\vspace{-0.2cm}
\caption{Chosen task groupings in Taskonomy. A task within a group is highlighted in bold when this task appears in multiple groups, and it is chosen to ``serve'' from this assigned task grouping. Human experts 1 and 2 are two industry professionals with an extensive background in computer vision. For groups selected by human experts, we chose to serve the task with during validation loss during co-training of the selected task groups.}
\label{table:taskonomy_groupings}
\vspace{-0.3cm}
\end{table*}

\subsection{Meta-World}
We use the five tasks: ``reach'', ``push'', ``press button top'', ``open window'', and ``close window'' from Meta-World~\citep{metaworld}. We use 6 fully-connected layers with 400 hidden units for both the policy and the critic with weights shared across all tasks. For each iteration, we collect 600 data points for each environment and train the policy and the critic for 600 steps with a batch size 128 per task. We use the soft actor critic (SAC)~\citep{haarnoja2018soft} as our RL algorithm and adopt the default hyperparameters used in the public repository of SAC ({\color{urlorange} \small \url{https://github.com/rail-berkeley/softlearning}} at hash 59f9ad357b11b973b9c64dcfeb7cf7b856bbae91). We compute the transference on the critic loss $$J_Q(\theta) = \mathbb{E}_{(\mathbf{s},\mathbf{a})\sim\mathcal{D}}\left[\frac{1}{2}(Q(\mathbf{s},\mathbf{a}) - \hat{Q}(\mathbf{s},\mathbf{a})^2)\right]\, ,$$ where $\mathbf{s}$ and $\mathbf{a}$ denote the state and action, $\hat{Q}$ denotes the target $Q$ network, $\mathcal{D}$ denotes the off-policy dataset collected by the agent, and $\theta$ denotes the parameter of the critic $Q$ network.

\label{app:landscape}
\begin{figure*}[t!]
\vspace{-0.2cm}
    \centering
    \includegraphics[width=1.0\linewidth]{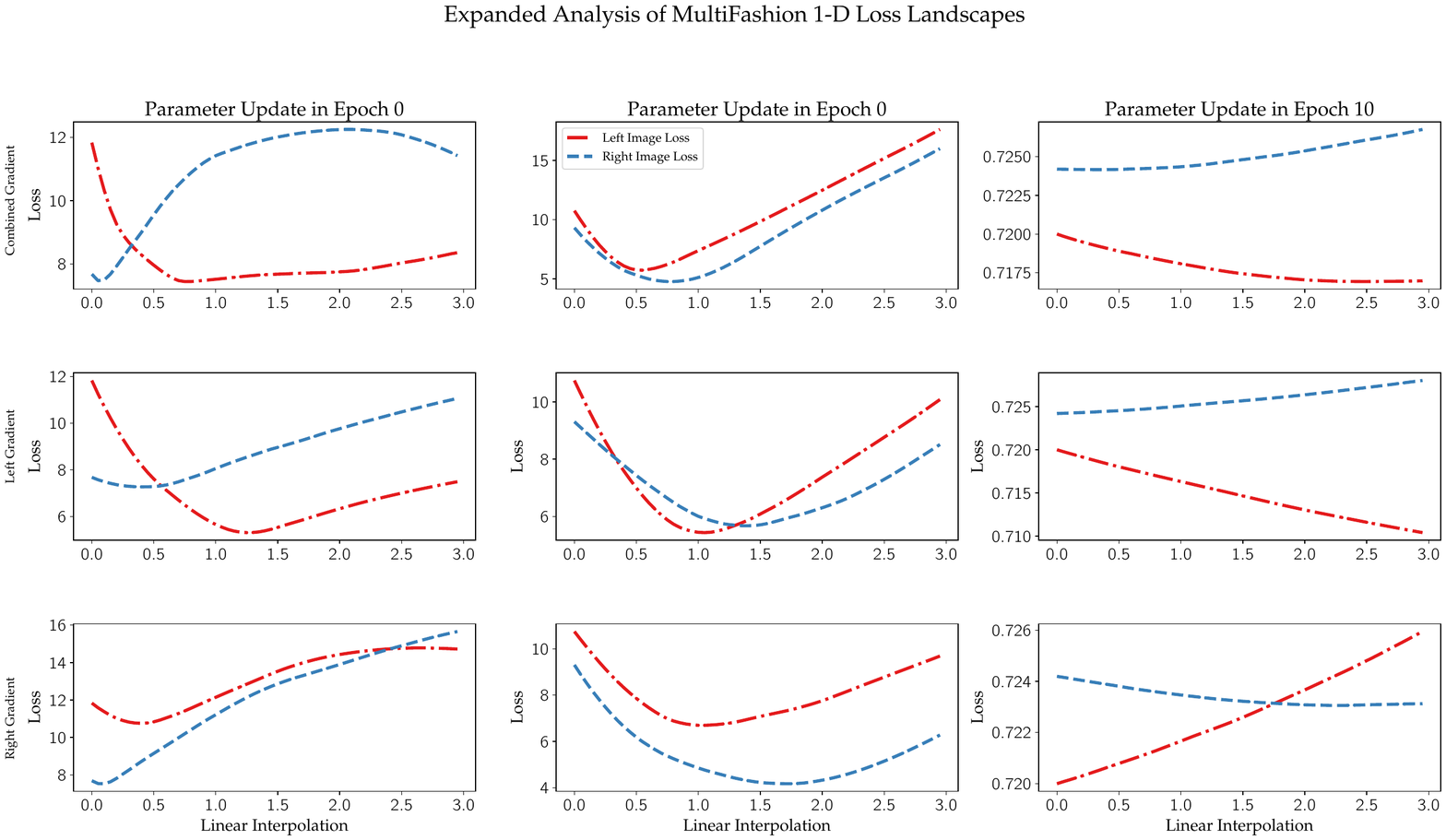}
    \vspace{-1.4cm}
    \caption{Expanded 1-d loss landscapes along each gradient direction.}
    \label{fig:expanded_1d_landscapes}
    \vspace{-0.3cm}
\end{figure*}

\subsection{MultiMNIST and MultiFashion}
\label{app:mnist}
Our experimental results on MultiMNIST and MultiFashion are generated using a combination of Keras and TensorFlow. We evaluate on the datasets released by \citet{pareto_moo} but further split $\frac{1}{6}$ of the training dataset into a validation set for final dataset splits of 100k/20k/20k train/valid/test.

The model architecture is loosely based on LeNet \citep{mnist}, and the model uses a momentum optimizer with momentum set to 0.9. The \textit{lookahead} loss is computed by simulating the full momentum update to the shared parameters rather than the SGD update described in Section~\ref{sec:measuring_transference}. The learning rate of the MTL baseline was selected on a validation dataset over $\{1e-4, 5e-4, 5e-3, 1e-2, 5e-2\}$ using a schedule which halves the learning rate every 30 epochs. A coarse grid search of the task-specific weights with left image weight = 1. - right image weight yielded left weight = right weight = 0.5. All experimental methods shared the hyperparameters used by the baseline, and experimental results are reported on the test set.

\subsection{NYUv2 and Cityscapes}
We clone the MTAN repository released by \citet{liu2019end} ({\color{urlorange} \small \url{https://github.com/lorenmt/mtan}} at hash b6504093ea6d646dccf501bbbb525a4d85db96ba). The optimization uses Adam~\citep{adam}, and the \textit{lookahead} loss is computed by simulating the full Adam update to the shared parameters rather than the SGD update described in Section~\ref{sec:measuring_transference}.

We run all MTAN experiments with the default hyperparameters specified in the github repository with the exception of reducing the number of steps to 100. We find significant overfitting begins after this stage and degrades the performance of the MTAN baseline.

\subsection{MovieLens}
We follow a similar experimental setup to~\citet{wang2020small}. In particular, we use the MovieLens 1M dataset~\citep{harper2015movielens} which records 1 million ratings from 6000 users on 4000 movies. We further augment the dataset with negative examples and sample 1.6 million samples from the augmented dataset such that each user has an equal number of watched and unwatched videos. Our augmented dataset will be released along with our code.

For every user-movie pairing, we create a binary prediction task to predict if the user has watched the movie as well as a regression task to predict the rating (1-5 stars) the user gave the movie. The regression task predicting rating is conditioned on click -- movie-user pairings in which the user did not watch the movie are zeroed out at training and during evaluation. 

We train with an Adam~\citep{adam} optimizer using typical parameters (beta1 = 0.9, beta2=0.999, epsilon=1e-8). We also use a learning rate schedule which lowers the learning rate every 25 epochs and train for 100 epochs in total. The model adapts the input with embeddings of the following dimensions: \{uid: 40, gender: 5, age: 10, occupation: 20, zip: 20, vid: 40, title: 20, genre: 20\}. The embeddings are concatenated together and fed through simple network consisting of \[Dense(50), Dropout(0.05), Dense(10), Dropout(0.01)\] with a final task-specific projection layer to adapt the shared representation to a watch or ranking prediction. To further decrease variability, the accuracy of the final 5 epochs are averaged together and reported in Table~\ref{table:movielens}.

\section{Expanded Geometric Analysis}
\Figref{fig:1dlosslandscape} was created by halting training in a given epoch immediately after either the left or the right task gradient update manifests higher transference than the $\sfrac{1}{2}(\text{left } + \text{ right})$ (i.e. combined) gradient update. We then applied the parameter update to the shared parameters using SGD with momentum to create a linear interpolation between the current parameter values and the parameter values following this update. We then extend this interpolation 3$\times$ past the typical update to measure the curvature along the direction of the update.

\Figref{fig:expanded_1d_landscapes} shows the loss along a linear interpolation of the left image gradient, the right image gradient, and the combined gradient directions. The columns coincide with the total loss plot presented in \Figref{fig:1dlosslandscape}. For instance, Column 1, Row 2 plots the left and right loss along the left gradient step which corresponds to the dashed red line representing the total loss along the left gradient step in \Figref{fig:1dlosslandscape}. Similarly, Column 2, Row 2 of \Figref{fig:expanded_1d_landscapes} plots the left and right loss along the right gradient step which sum together to create the blue dashed line in the center plot of \Figref{fig:1dlosslandscape}.

In \Figref{fig:2dlosslandscape}, we plot the 2-dimensional loss landscape of the left and right loss as well as the combined loss for MultiFashion. To generate the plots, we first sample two random directions in the parameter space and then scale the norms of these directions to be equal to the norm of the parameters. Next, we interpolate the parameters along these two directions in the range $[-0.1, +0.1]$ times the norm of the parameters.

The left image loss depicts a smooth landscape whereas the right image loss is highly non-smooth. Notice that the level sets of the combined  (i.e. average) loss is higher than those of the left loss. For this step, {\color{modelblue} \method} chooses the left gradient for the shared parameter update which aligns with the curvature discrepancy between the right image loss and the left image loss. 




\end{document}